\documentclass{article}

\usepackage[preprint]{neurips_2025}

\usepackage[utf8]{inputenc} 
\usepackage[T1]{fontenc}    
\usepackage{hyperref}       
\usepackage{url}            
\usepackage{booktabs}       
\usepackage{amsfonts}       
\usepackage{nicefrac}       
\usepackage{microtype}      
\usepackage[table]{xcolor}
\usepackage{multirow}
\usepackage{subcaption}
\usepackage{amsmath}
\usepackage{graphicx}
\usepackage{tcolorbox}
\usepackage{CJKutf8}

\title{GenCLS++: Pushing the Boundaries of Generative Classification in LLMs Through Comprehensive SFT and RL Studies Across Diverse Datasets}

\author{
 Mingqian He\textsuperscript{\normalfont 1},
 Fei Zhao\textsuperscript{\normalfont 2\textdagger},
 Chonggang Lu\textsuperscript{\normalfont 2},
 Ziyan Liu\textsuperscript{\normalfont 3},\\
 \textbf{Yue Wang}\textsuperscript{4},
 \textbf{Haofu Qian}\textsuperscript{1}\\
 \textsuperscript{1}Zhejiang University, \textsuperscript{2}Xiaohongshu Inc., \\
 \textsuperscript{3}Beijing University of Posts and Telecommunications, \textsuperscript{4}Nanjing University \\
 \texttt{mingqianhe@zju.edu.cn}, \texttt{zhaofei5@xiaohongshu.com}
}

\begin{document}
\renewcommand{\thefootnote}{\textdagger}
\footnotetext{Corresponding author.}

\maketitle

\begin{abstract}

 As a fundamental task in machine learning, text classification plays a crucial role in many areas. With the rapid scaling of Large Language Models (LLMs), particularly through reinforcement learning (RL), there is a growing need for more capable discriminators. Consequently, advances in classification are becoming increasingly vital for enhancing the overall capabilities of LLMs. Traditional discriminative methods map text to labels but overlook LLMs' intrinsic generative strengths. Generative classification addresses this by prompting the model to directly output labels. However, existing studies still rely on simple SFT alone, seldom probing the interplay between training and inference prompts, and no work has systematically leveraged RL for generative text classifiers and unified SFT, RL, and inference-time prompting in one framework. We bridge this gap with \textbf{GenCLS++}, a framework that jointly optimizes SFT and RL while systematically exploring five high-level strategy dimensions—in-context learning variants, category definitions, explicit uncertainty labels, semantically irrelevant numeric labels, and perplexity-based decoding—during both training and inference. After an SFT “policy warm-up,” we apply RL with a simple rule-based reward, yielding sizable extra gains. \textbf{Across seven datasets, GenCLS++ achieves an average accuracy improvement of 3.46\% relative to the naive SFT baseline; on public datasets, this improvement rises to 4.00\%.} Notably, unlike reasoning-intensive tasks that benefit from explicit thinking processes, we find that classification tasks perform better without such reasoning steps. These insights into the role of explicit reasoning provide valuable guidance for future LLM applications.
 
\end{abstract}

\section{Introduction}
With the rapid advancement of Large Language Models (LLMs) \citep{anthropic2024claude, google2024gemini, openai2024gpt4o}, remarkable progress has been achieved in enhancing their generative capabilities, particularly in the domain of reasoning. Throughout this development, well-designed discriminators play a crucial role, whether in aligning model outputs with human preferences \citep{schulman2017proximal, ziegler2019fine, ouyang2022training} or scaling model capabilities through effective reward signals \citep{guo2025deepseek, yu2025dapo, bytedance2025seed}. The emergence of DeepSeek-R1 \citep{guo2025deepseek} highlights the effectiveness of rule-based rewards in domains such as mathematics and code. However, in broader scenarios where golden answers are not readily available, learned discriminators remain indispensable for providing reliable reward signals \citep{bytedance2025seed, liu2025inference}.

Building on this insight, we explore methods to enhance the performance of discriminator models by focusing on the closely related task of classification. Traditional discriminative approaches \citep{ruan2024large, muennighoff2022mteb, cobbe2021training, yu2023ovm} typically involve using a randomly initialized value head with a pre-trained language model to map text to labels, relying on the representation token to predict class probabilities. Although this method is widely adopted, it introduces an inherent mismatch between the randomly initialized value head and the carefully optimized language model, potentially leading to suboptimal performance \citep{zhang2024generative, ye2024beyond}. This discrepancy may hinder the model from fully exploiting the generative capabilities already embedded within LLMs.

Recent advancements in prompt-based learning offer an alternative by guiding LLMs to perform classification directly through language generation \citep{parikh2023exploring, rouzegar2024enhancing}. This generative approach naturally aligns with the intrinsic training paradigm of LLMs, effectively leveraging their native language understanding and generation capabilities. Compared to traditional methods, this approach offers several advantages:
\begin{itemize}
    \item Benefits from LLM Improvement: Generative methods enable classification tasks to directly benefit from ongoing advancements in LLM capabilities, naturally scaling classification accuracy with improved underlying LLM performance.
    \item Greater Flexibility: Generative methods allow the addition of new categories without extensive training or altering the model architecture. Traditional methods necessitate adjusting dimensions and retraining when new labels are introduced.
\end{itemize}


Despite its intuitive appeal, most methods adopt a simple, and identical prompt strategy for both training and inference. The systematic exploration of diverse prompt strategies for both stages remains limited, with the effects of using different prompts during these stages not yet thoroughly investigated or quantified. To address this gap, we propose \textbf{GenCLS++}, a prompt-based generative classification framework that systematically explores five high-level strategy dimensions: In-Context Learning (ICL) variants (semantic retrieval vs. fixed exemplars, and varying shot counts), category definitions, explicit uncertainty labels, semantically irrelevant labels, and perplexity-based decoding, during both supervised fine-tuning and inference. Furthermore, inspired by recent advances, we integrate reinforcement learning (RL) into GenCLS++, resulting in additional performance gains and underscoring the potential of unifying supervised and reinforcement learning paradigms. We empirically evaluate GenCLS++ across seven diverse datasets, comprising both publicly available and internal data. Our findings reveal that GenCLS++ significantly enhances classification accuracy, achieving an average accuracy improvement of 3.46\% relative to the commonly used naive SFT baseline. This improvement rises to 4.00\% on public datasets.

Interestingly, our results challenge assumptions derived from related reasoning-intensive tasks. While explicit reasoning steps have shown significant performance improvements in such tasks, we find that classification tasks often achieve optimal results without explicit reasoning prompts, consistent with some recent studies \citep{li2025cls}. These findings offer new insights into the role and necessity of explicit reasoning in classification contexts.

Our contributions can be summarized as follows:
\begin{itemize}    
    \item We conduct a comprehensive analysis of a wide range of prompt strategies for classification tasks. Our findings reveal that specific combinations can significantly outperform the naive SFT approach, highlighting their effectiveness in enhancing model performance.
    \item We integrate RL to further boost performance. Our experiments indicate that supervised fine-tuning for warm-up initialization delivers a significant relative improvement in accuracy, with an average gain of 18.18\% compared to training directly from the base model.
    \item Motivated by recent investigations into reasoning-based inference and the observation that RL tends to produce much shorter responses, we find that models achieve better performance on classification tasks by directly generating answers without explicit reasoning steps.
\end{itemize}

\section{Related Work}
\paragraph{LLMs for Classification}
Compared to the traditional discriminative approach of employing a value head to map text to labels, recent studies have explored a generative strategy in which LLMs perform classification through prompt engineering \citep{qin2023chatgpt, sun2023text, peskine2023definitions, milios2023context}, augmenting the prompt with few‑shot examples and category definitions. However, few studies have taken the next step of fine‑tuning LLMs to generate class labels \citep{parikh2023exploring}, and several reports indicate that the generative approach underperforms on certain classification benchmarks \citep{ruan2024large}. Our study advances prior work by systematically examining various combinations of training-time and inference‑time strategies. With this framework we achieve consistently higher accuracy across these datasets, providing strong evidence of the generative paradigm’s potential for classification tasks.
\paragraph{RL for LLM Training}
Reinforcement learning (RL) now plays a pivotal role in training LLMs. It is used not only to align outputs with human preferences through Reinforcement Learning from Human Feedback (RLHF) \citep{ziegler2019fine, ouyang2022training}, but also to enhance models’ reasoning abilities, as recently demonstrated by DeepSeek‑R1 \citep{guo2025deepseek}. These applications underscore RL’s vast potential to drive further advancements in LLMs. In this paper, we investigate how RL can improve performance on classification tasks and present several noteworthy empirical findings. Given that PPO \citep{schulman2017proximal} incurs substantial computational overhead and rule‑based rewards have shown to be effective, we conduct most of our experiments using the more efficient Reinforce++ algorithm \citep{hu2025reinforce++}, along with carefully designed rule‑based reward functions.

\begin{figure*}[!ht] 
    \centering
    \includegraphics[width=\linewidth]{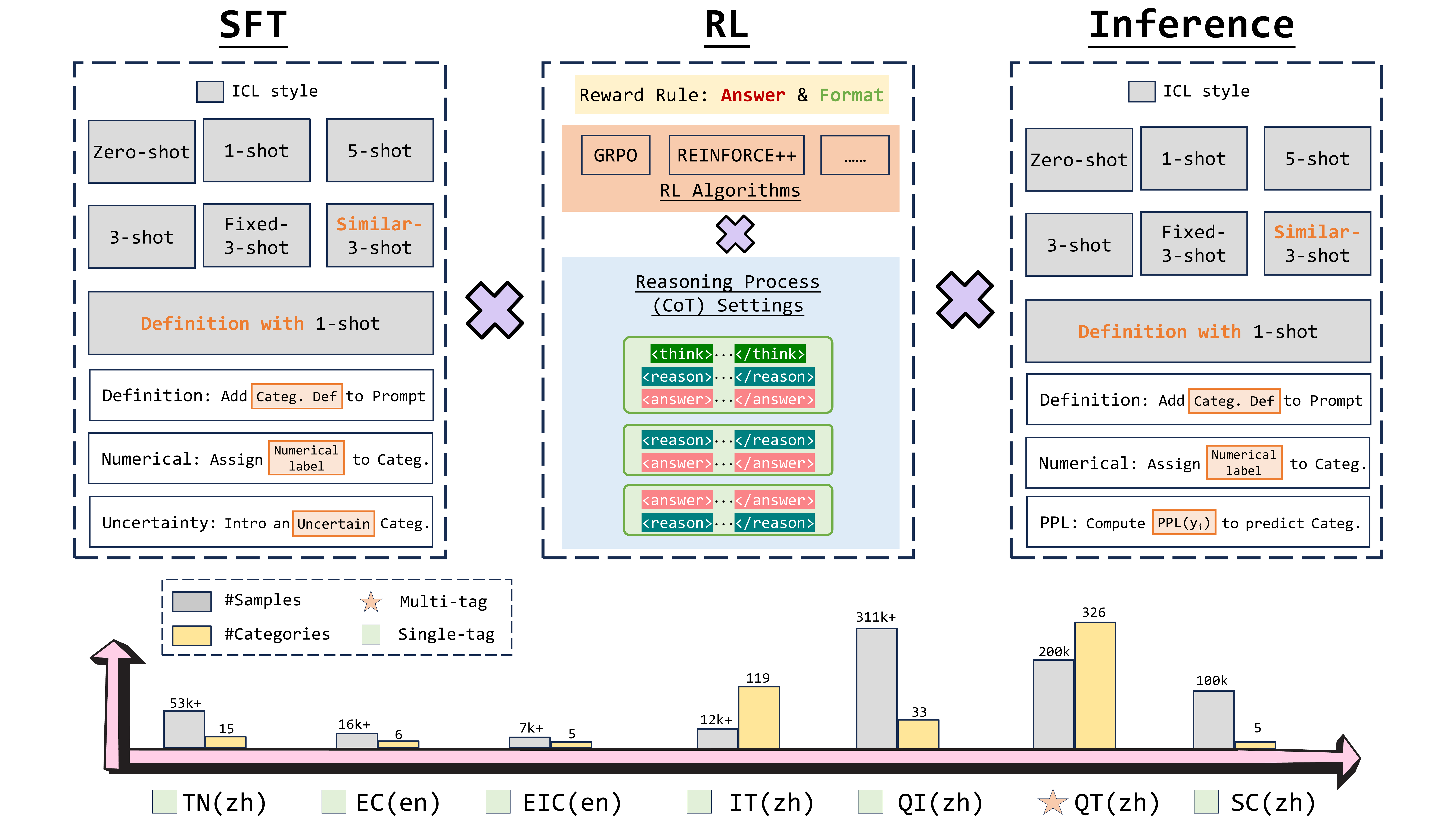}
    \caption{An overview of the GenCLS++ framework. It explores diverse combinations of training and inference strategies for classification tasks and incorporates RL to further enhance performance. We conduct comprehensive experiments on seven datasets, encompassing different languages, varying numbers of categories, and diverse data types.}
    \label{fig:method}
\end{figure*}

\section{Method}
Traditional LLM-based classifiers do not fully utilize the text generation capabilities of pretrained LLMs. To address this issue, we propose training generative classifiers using standard next token prediction. Specifically, instead of obtaining each category’s probability through a representative token, the language model predicts categories using its own probability distribution over tokens. This approach preserves the model’s generative abilities, since classification is merely another token prediction, while also offering several advantages that naturally arise from LLMs, such as a unified paradigm for pretraining and classification, and the ability to scale inference time compute. The overview of our method is shown in Figure~\ref{fig:method}.
\subsection{Exploring Different Strategy Combinations in SFT and Inference}
\label{3.1}
Let $\mathbf{x}$ denote the input to be classified. A generative classifier $\pi_\theta$ predicts the gold category $\mathbf{y_{\text{gold}}}$ using tokens. This is achieved by maximizing $\log \pi_\theta\bigl(\mathbf{y_{\text{gold}}} \mid (\mathbf{p}, \mathbf{x})\bigr)$, where $\mathbf{p}$ represents a particular prompt strategy from the strategy pool $\mathcal{P}$. To do so, we minimize the supervised fine-tuning (SFT) loss on the dataset $\mathcal{D}$, which contains input–class pairs:
$$
\mathcal{L}_{\text{SFT}}(\theta,\mathcal{D})
= - \mathbb{E}_{(\mathbf{x},\mathbf{y}) \sim \mathcal{D}}\Biggl[\sum_{t=1}^{|\mathbf{y}|} \log \pi_\theta\bigl(y_t \mid \mathbf{p}, \mathbf{x}, \mathbf{y}_{<t}\bigr)\Biggr].
$$
We aim to explore how different prompt strategies, applied during both training and inference, affect classification performance within a purely generative paradigm. Below, we describe the types of prompt strategies in $\mathcal{P}$ that we employed; examples of each type are provided in Appendix~\ref{A}.

\paragraph{Zero-shot}
The model receives only a general task description without labeled examples or detailed definitions:
$$\mathcal{P}_{\text{zero-shot}} = P_{\text{zero-shot}}(\mathbf{x})$$
where $P_{\text{zero-shot}}$ is a prompt template that introduces the classification task and requests a prediction for the input $\mathbf{x}$ without providing examples.

\paragraph{N-shot}
We include $N$ labeled examples (randomly selected from the training set) in the prompt $\bigl(N \in \{1,3,5\}\bigr)$, providing the model with exemplar input–label pairs to guide classification:
$$\mathcal{P}_{N\text{-shot}} = P_{N\text{-shot}}((\mathbf{x}_1, \mathbf{y}_1), (\mathbf{x}_2, \mathbf{y}_2), \ldots, (\mathbf{x}_N, \mathbf{y}_N), \mathbf{x})$$
where $(\mathbf{x}_i, \mathbf{y}_i)$ are randomly sampled examples from the training set, and $P_{N\text{-shot}}$ is a prompt template that includes these examples followed by the input $\mathbf{x}$ to be classified.

\paragraph{Fixed-3-shot}
The same three labeled examples appear in the prompt for every test case:
$$\mathcal{P}_{\text{fixed-3}} = P_{\text{fixed-3}}((\mathbf{x}_1^{\text{fixed}}, \mathbf{y}_1^{\text{fixed}}), (\mathbf{x}_2^{\text{fixed}}, \mathbf{y}_2^{\text{fixed}}), (\mathbf{x}_3^{\text{fixed}}, \mathbf{y}_3^{\text{fixed}}), \mathbf{x})$$
where $\{(\mathbf{x}_i^{\text{fixed}}, \mathbf{y}_i^{\text{fixed}})\}_{i=1}^3$ are three fixed examples selected once from the training set, and $P_{\text{fixed-3}}$ is a prompt template that includes these fixed examples followed by the input $\mathbf{x}$ to be classified.

\paragraph{Similar-3-shot}
We retrieve the three training examples most similar to the new input and include them in the prompt:
$$\mathcal{P}_{\text{similar-3}} = P_{\text{similar-3}}((\mathbf{x}_1^{\text{sim}}(\mathbf{x}), \mathbf{y}_1^{\text{sim}}(\mathbf{x})), (\mathbf{x}_2^{\text{sim}}(\mathbf{x}), \mathbf{y}_2^{\text{sim}}(\mathbf{x})), (\mathbf{x}_3^{\text{sim}}(\mathbf{x}), \mathbf{y}_3^{\text{sim}}(\mathbf{x})), \mathbf{x})$$
where $\{(\mathbf{x}_i^{\text{sim}}(\mathbf{x}), \mathbf{y}_i^{\text{sim}}(\mathbf{x}))\}_{i=1}^3$ are the three examples from the training set that have the highest cosine similarity to $\mathbf{x}$ in the embedding space, and $P_{\text{similar-3}}$ is a prompt template that includes these similar examples followed by the input $\mathbf{x}$ to be classified. Specifically, we use a text encoder to generate embeddings and compute the similarity:
$$\{(\mathbf{x}_i^{\text{sim}}(\mathbf{x}), \mathbf{y}_i^{\text{sim}}(\mathbf{x}))\}_{i=1}^3 = \text{Top3}_{(\mathbf{x}', \mathbf{y}') \in \mathcal{D}_{\text{train}}} \cos(E(\mathbf{x}), E(\mathbf{x}'))$$
where $E(\cdot)$ is the embedding function of the text encoder and $\cos(\cdot,\cdot)$ denotes cosine similarity.

\paragraph{Definition}
We prepend concise text definitions of each target category to the prompt:
$$\mathcal{P}_{\text{def}} = P_{\text{def}}(\mathcal{T}_1, \mathcal{T}_2, \ldots, \mathcal{T}_C, \mathbf{x})$$
where $\mathcal{T}_c$ is the textual definition of category $c$, $C$ is the total number of categories, and $P_{\text{def}}$ is a prompt template that includes these definitions followed by the input $\mathbf{x}$ to be classified. These definitions are generated by prompting a LLM to provide explanations for each class label in the dataset. For example, for an emotion classification task with label "anger", the generated definition is: "anger: contains strong negative feelings like anger, annoyance, indignation, involving injustice, conflict, frustration, etc." These generated definitions are then incorporated into the model's prompt to provide clear semantic meanings of the categories before classification. For our implementation, we use ``GPT-4o-2024-11-20'' as the LLM to generate these definitions.

\paragraph{Definition with 1-shot}
In addition to including category definitions in the prompt, we also include a single labeled example:
$$\mathcal{P}_{\text{def+1}} = P_{\text{def+1}}(\mathcal{T}_1, \mathcal{T}_2, \ldots, \mathcal{T}_C, (\mathbf{x}_{\text{ex}}, \mathbf{y}_{\text{ex}}), \mathbf{x})$$
where $(\mathbf{x}_{\text{ex}}, \mathbf{y}_{\text{ex}})$ is an example from $\mathcal{D}$, and $P_{\text{def+1}}$ is a prompt template that includes both category definitions and the example, followed by the input $\mathbf{x}$ to be classified.

\paragraph{Numerical (semantically irrelevant labels)}
We assign each category a numerical label and prompt the model to output the corresponding number. This approach is non-semantic, relying purely on arbitrary number assignments rather than meaningful category descriptions. Formally,
$$\mathcal{P}_{\text{num}} = P_{\text{num}}((\mathbf{c}_1, 1), (\mathbf{c}_2, 2), \ldots, (\mathbf{c}_C, C), \mathbf{x})$$
Here, $\mathbf{c}_i$ denotes the name of category $i$, and $P_{\mathrm{num}}$ is a prompt template that associates each category with its numerical label, instructing the model to output only the number corresponding to the predicted category for a given input $\mathbf{x}$.

\paragraph{Uncertainty}
We introduce a class called "Uncertain" for ambiguous training examples. To identify these examples, we use the two best-performing models (based on zero-shot accuracy) trained on the original dataset. For each training example $(\mathbf{x}_i, \mathbf{y}_i) \in \mathcal{D}$, we redefine the label as:
$$\hat{\mathbf{y}}_i = \begin{cases}
\text{``Uncertain''} & \text{if } M_1(\mathbf{x}_i) \neq \mathbf{y}_i \text{ and } M_2(\mathbf{x}_i) \neq \mathbf{y}_i \\
\mathbf{y}_i & \text{otherwise}
\end{cases}$$
where $M_1(\mathbf{x}_i)$ and $M_2(\mathbf{x}_i)$ are the predictions of the two models in the zero-shot setting. We limit the proportion of examples labeled as "Uncertain" to at most 10\% of the training data. If more examples qualify as uncertain, we select the 10\% with the lowest average prediction confidence across both models. This creates a modified dataset $\hat{\mathcal{D}} = \{(\mathbf{x}_i, \hat{\mathbf{y}}_i)\}_{i=1}^{|\mathcal{D}|}$ that is used to train a new model:
$$\mathcal{P}_{\text{uncertainty}} = P_{\text{uncertainty}}(\mathbf{x})$$
where $P_{\text{uncertainty}}$ is a prompt template similar to the zero-shot prompt but designed to accommodate the additional "Uncertain" class during training. During inference, this model is constrained to predict only from the original class set, excluding the "Uncertain" class.

\paragraph{Perplexity}
For each candidate class $\mathbf{y}_i$, we append it to the input and compute its perplexity $\mathrm{PPL}(\mathbf{y}_i)$ as follows:
$$
\mathrm{PPL}(\mathbf{y}_i) = \exp\Bigl\{ -\frac{1}{|\mathbf{y}_i|} \sum_{t=1}^{|\mathbf{y}_i|} \log \pi_{\theta}\bigl(y_{t}\mid \mathcal{P}_{\text{base}}(\mathbf{x}), \mathbf{y}_{<t} \bigr) \Bigr\}
$$
where $\mathcal{P}_{\text{base}}(\mathbf{x})$ is a base prompt. We then select the class with the lowest perplexity as our prediction:
$$\hat{y} = \arg\min_{\mathbf{y}_i \in \mathcal{C}} \mathrm{PPL}(\mathbf{y}_i)$$
Note that this strategy is employed only at inference time.

We apply various strategies to train the model and subsequently evaluate each resulting model with different prompt strategies (e.g., trained with definitions, evaluated in a zero-shot setting), as illustrated in Figure~\ref{fig:method}. In contrast to traditional few-shot learning, which uses the same prompt type for both training and inference, our approach enables a more fine-grained analysis of how different strategies affect performance at each stage.

\subsection{Reinforcement Learning}
Building on the success of DeepSeek‑R1 \citep{guo2025deepseek}, which shows that reinforcement learning (RL) can markedly enhance the reasoning ability of language models, we explore RL for generative classification. Specifically, we fine‑tune our model with a rule‑based reward function to gauge the effectiveness of RL in this setting.
\subsubsection{Policy Warm-up}
During the warm-up phase, we equip the policy model with foundational classification capabilities by performing supervised fine-tuning on the dataset $\mathcal{D}$. We find that this phase has a significant impact on the subsequent performance of RL. Furthermore, we investigate how different start models affect the final performance. Detailed results and discussions are presented in Sections~\ref{5.1} and~\ref{5.2}.
\subsubsection{RL with Reasoning}
\paragraph{System Prompt} We first follow \citet{guo2025deepseek}’s paradigm, encouraging models to engage in a reasoning (thinking) process before producing the final answer. The prompt is defined as follows: "Please output your answer in the format: <reason> reasoning process here </reason> <answer> answer here </answer>."
\paragraph{Reward Function} Similarly, we design a two-part rule-based reward function for RL: format reward and accuracy reward. The format reward verifies that the response follows the required structured format, ensuring that every part appears in the correct order and is enclosed in the appropriate tags:
$$
R_{\text{format}} =
\begin{cases}
1, & \text{if the format is correct},\\
0, & \text{otherwise}.
\end{cases}
$$

The accuracy reward measures whether the model’s prediction matches the gold label $\mathbf{y_{\text{gold}}}$:
$$
R_{\text{accuracy}} =
\begin{cases}
1, & \text{if } y = \mathbf{y_{\text{gold}}},\\
0, & \text{otherwise}.
\end{cases}
$$
The final reward function $R$ is a combination of the two rewards and is defined as: 
$$R = R_{\text{format}} + R_{\text{accuracy}}$$

\subsubsection{RL without Reasoning}
Unlike reasoning-driven tasks such as mathematics and code generation, we observed that during the reinforcement learning process in classification tasks, the response length fluctuates and may even decrease rapidly. Additionally, comparative experiments revealed that the inclusion of rationale does not seem to contribute to performance improvement, as discussed in Section~\ref{5.2}. This phenomenon has also been observed in other tasks, such as commonsense question answering \citep{jiang2025mme,sprague2024cot,sui2025stop} and vision classification \citep{li2025cls}. These findings suggest that chain-of-thought (CoT) reasoning may not be essential for all tasks. Motivated by these insights, we investigate RL in classification tasks without incorporating a reasoning process.

\paragraph{System Prompt} Unlike most current RL-based scaling methods, which encourage models to repeatedly reason and verify, the prompt in our method directs the model to output the classification result directly, e.g., "Please output your answer."

\paragraph{Reward Function} Since we no longer need to distinguish between reasoning and the answer, we eliminate the need for conventional format rewards. Instead, we solely use an accuracy reward, which checks whether the model’s output matches the ground truth exactly. It is defined as follows:
$$
R_{\text{accuracy}} =
\begin{cases}
1, & \text{if } y = \mathbf{y_{\text{gold}}}, \\
0, & \text{otherwise}.
\end{cases}
$$
We adopt Reinforce++ \citep{hu2025reinforce++} as our reinforcement learning algorithm. In Section~\ref{5.3}, we compare it with several widely used baselines, such as Reinforce++ Baseline and GRPO \citep{shao2024deepseekmath}, and find that Reinforce++ consistently delivers higher accuracy while requiring less training time, demonstrating advantages in both performance and efficiency.

\begin{table}[ht]
    \centering
    \caption{Experimental Results (Accuracy \& macro-F1 Score, \%). \colorbox{gray!30}{Gray} indicates that the training strategy is not aligned with the best inference strategy. \textbf{Bold} indicates the best result, and \underline{underline} indicates the second best. * reported by \citet{ruan2024large}; \textdagger reported by \citet{xu2020clue}.}
    \label{tab:results}
    \begin{subtable}[t]{\textwidth}
        \small
        \centering
        \caption{Part 1: Results on Public Datasets}
        \begin{tabular}{lcccccccc}
        \toprule
        & \multicolumn{2}{c}{EC} & \multicolumn{2}{c}{EIC} & \multicolumn{2}{c}{IFLYTEK} & \multicolumn{2}{c}{TNEWS} \\
        \cmidrule(lr){2-3}\cmidrule(lr){4-5}\cmidrule(lr){6-7}\cmidrule(lr){8-9}
        \textbf{Training Method} &Acc. &macro-F1  &Acc. &macro-F1  &Acc. &macro-F1  &Acc. &macro-F1 \\
        \midrule
        Base model             &68.75 &33.94  &55.02 &49.85  &57.41 &25.93  &58.13 &22.92 \\
        Discriminative method  &94.10\textsuperscript{*} &89.60\textsuperscript{*}  &84.40\textsuperscript{*} &82.20\textsuperscript{*}  &62.98\textsuperscript{\textdagger} &-      &59.46\textsuperscript{\textdagger} &- \\
        Naive SFT              &93.70 &90.17  &82.74 &81.73  &61.18 &45.42  &60.83 &59.39 \\
        \midrule
        Zero-shot              &\cellcolor{gray!30}93.75 &\cellcolor{gray!30}\underline{90.31}  &\cellcolor{gray!30}84.04 &\cellcolor{gray!30}82.93  &\cellcolor{gray!30}62.83 &\cellcolor{gray!30}46.69  &\cellcolor{gray!30}62.12 &\cellcolor{gray!30}59.16 \\
        1-shot                 &\cellcolor{gray!30}93.15 &\cellcolor{gray!30}89.45  &\cellcolor{gray!30}83.82 &\cellcolor{gray!30}82.90  &\cellcolor{gray!30}63.33 &\cellcolor{gray!30}47.15  &\cellcolor{gray!30}61.98 &\cellcolor{gray!30}60.49 \\
        3-shot                 &\cellcolor{gray!30}93.45 &\cellcolor{gray!30}89.13  &8\cellcolor{gray!30}5.03 &\cellcolor{gray!30}\underline{83.75}  &\cellcolor{gray!30}62.91 &\cellcolor{gray!30}\underline{48.51}  &\cellcolor{gray!30}62.06 &\cellcolor{gray!30}60.53 \\
        5-shot                 &\underline{94.15} &89.83  &\cellcolor{gray!30}84.13 &\cellcolor{gray!30}83.39  &\cellcolor{gray!30}62.52 &\cellcolor{gray!30}44.93  &\cellcolor{gray!30}62.54 &\cellcolor{gray!30}58.41 \\
        Fixed-3-shot           &93.80 &89.92  &\cellcolor{gray!30}83.17 &\cellcolor{gray!30}82.49  &63.52 &45.26  &\cellcolor{gray!30}62.25 &\cellcolor{gray!30}58.94 \\
        Similar-3-shot         &93.90 &89.44  &\cellcolor{gray!30}82.18 &\cellcolor{gray!30}79.35  &\cellcolor{gray!30}62.83 &\cellcolor{gray!30}47.69  &\underline{63.30} &\underline{61.31} \\
        Definition             &\cellcolor{gray!30}93.30 &\cellcolor{gray!30}88.31  &83.26 &82.30  &\cellcolor{gray!30}63.64 &\cellcolor{gray!30}44.41  &\cellcolor{gray!30}61.37 &\cellcolor{gray!30}59.26 \\
        Definition with 1-shot &\cellcolor{gray!30}93.80 &\cellcolor{gray!30}89.70  &\cellcolor{gray!30}84.34 &\cellcolor{gray!30}82.79  &\cellcolor{gray!30}63.37 &\cellcolor{gray!30}47.47  &\cellcolor{gray!30}62.20 &\cellcolor{gray!30}60.65 \\
        Numerical              &93.65 &89.93  &83.17 &81.89  &62.29 &46.29  &61.24 &57.09 \\
        Uncertainty            &93.55 &90.01  &\underline{85.08} &83.74  &\cellcolor{gray!30}\underline{63.76} &\cellcolor{gray!30}47.85  &\cellcolor{gray!30}61.97 &\cellcolor{gray!30}58.15 \\
        \midrule
        GenCLS++ (RL)                      &\bf{94.50} &\bf{90.57}  &\bf{85.86} &\bf{84.72}  &\bf{64.91} &\bf{49.27}  &\bf{64.04} &\bf{62.35} \\
        \bottomrule
        \end{tabular}
    \end{subtable}

    \begin{subtable}[t]{\textwidth}
        \small
        \centering
        \caption{Part 2: Results on Proprietary Datasets}
        \begin{tabular}{lcccccc}
        \toprule
        & \multicolumn{2}{c}{Query Intent} & \multicolumn{2}{c}{Search Correlation} & \multicolumn{2}{c}{Query Taxonomy} \\
        \cmidrule(lr){2-3}\cmidrule(lr){4-5}\cmidrule(lr){6-7}
        \textbf{Training Method}  &Acc. &macro-F1  &Acc. &macro-F1  &Acc. &macro-F1 \\
        \midrule
        Base model                &74.91 &18.59  &41.91 &34.24  &26.51 &14.70 \\
        Naive SFT           &92.28 &86.33  &67.43 &58.64  &51.43 &43.10 \\
        \midrule
        Zero-shot                 &\cellcolor{gray!30}92.30 &\cellcolor{gray!30}86.27  &65.27 &54.44  &\cellcolor{gray!30}53.25 &\cellcolor{gray!30}43.13 \\
        1-shot                    &92.48 &\bf{87.34}  &67.37 &59.10  &\cellcolor{gray!30}53.09 &\cellcolor{gray!30}44.02 \\
        3-shot                    &\cellcolor{gray!30}92.44 &\cellcolor{gray!30}86.55  &\cellcolor{gray!30}67.53 &\cellcolor{gray!30}59.23  &\cellcolor{gray!30}53.38 &\cellcolor{gray!30}43.84 \\
        5-shot                    &\cellcolor{gray!30}\underline{92.52} &\cellcolor{gray!30}\underline{86.95}  &66.73 &56.42  &\cellcolor{gray!30}53.95 &\cellcolor{gray!30}\underline{44.52} \\
        Fixed-3-shot              &92.36 &86.17  &\cellcolor{gray!30}64.76 &\cellcolor{gray!30}50.08  &\cellcolor{gray!30}\underline{54.03} &\cellcolor{gray!30}43.60 \\
        Similar-3-shot            &92.22 &86.40  &67.63 &60.01  &52.99 &44.11 \\
        Definition                &92.23 &85.91  &\underline{68.60} &\underline{62.25}  &-     &-     \\
        Definition with 1-shot    &\cellcolor{gray!30}92.41 &\cellcolor{gray!30}78.02  &\cellcolor{gray!30}67.40 &\cellcolor{gray!30}58.51  &- &- \\
        Numerical                 &92.52 &86.34  &64.40 &47.89  &51.26 &42.87 \\
        Uncertainty               &92.36 &86.51  &65.57 &53.18  &50.21     &38.35  \\
        \midrule
        GenCLS++ (RL)                        &\bf{92.62} &86.86 &\bf{68.94} &\bf{65.08}  &\bf{54.31} &\bf{46.18} \\
        \bottomrule
        \end{tabular}
    \end{subtable}%
\end{table}

\begin{table*}[!ht]
\caption{Experimental results on the EIC dataset (Accuracy \& macro-F1 Score, \%). \colorbox{blue!30}{Blue} highlights the best inference strategy for each training method, while \textbf{bold} denotes the overall best performance across all settings.}
\label{tab:EIC}
\centering
\scriptsize
\setlength{\tabcolsep}{0.5pt}
\begin{tabular}{@{}l|ccccc|ccccc|ccccc@{}}
\toprule
\multirow{3}{*}{\textbf{Method}} & \multicolumn{5}{c|}{\textbf{1-shot}} & \multicolumn{5}{c|}{\textbf{3-shot}} & \multicolumn{5}{c}{\textbf{fix\_3\_shot}} \\
\cmidrule(lr){2-6} \cmidrule(lr){7-11} \cmidrule(lr){12-16}
 & fmt-suc & fmt-suc & fmt-suc & overall & overall & fmt-suc & fmt-suc & fmt-suc & overall & overall & fmt-suc & fmt-suc & fmt-suc & overall & overall \\
 & ratio & acc & macro-f1 & acc & macro-f1 & ratio & acc & macro-f1 & acc & macro-f1 & ratio & acc & macro-f1 & acc & macro-f1 \\
\midrule
Zero-shot & 100 & 83.43 & 82.40 & 83.43 & 82.40 & 100 & 83.39 & 82.67 & 83.39 & 82.67 & 100 & 82.74 & 82.22 & 82.74 & 82.22 \\
1-shot & 100 & 83.56 & 82.86 & 83.56 & 82.86 & 100 & \colorbox{blue!30}{83.82} & \colorbox{blue!30}{82.90} & \colorbox{blue!30}{83.82} & \colorbox{blue!30}{82.90} & 100 & 82.74 & 81.95 & 82.74 & 81.95 \\
3-shot & 100 & 84.08 & 82.88 & 84.08 & 82.88 & 100 & 84.60 & 83.28 & 84.60 & 83.28 & 100 & \colorbox{blue!30}{85.03} & 83.75 & \colorbox{blue!30}{85.03} & 83.75 \\
5-shot & 100 & 83.82 & 82.91 & 83.82 & 82.91 & 100 & \colorbox{blue!30}{84.13} & \colorbox{blue!30}{83.39} & \colorbox{blue!30}{84.13} & \colorbox{blue!30}{83.39} & 100 & 83.39 & 82.61 & 83.39 & 82.61 \\
Definition & 100 & 83.09 & 82.15 & 83.09 & 82.15 & 100 & 83.09 & 82.11 & 83.09 & 82.11 & 100 & 82.27 & 81.25 & 82.27 & 81.25 \\
Numerical & 100 & 38.41 & 4.83 & 38.41 & 4.83 & 100 & 48.18 & 6.61 & 48.18 & 6.61 & 100 & 50.56 & 7.69 & 50.56 & 7.69 \\
Similar-3-shot & 100 & 81.19 & 78.28 & 81.19 & 78.28 & 100 & 81.88 & 79.26 & 81.88 & 79.26 & 100 & \colorbox{blue!30}{82.18} & 79.35 & \colorbox{blue!30}{82.18} & 79.35 \\
Fixed-3-shot & 100 & 81.75 & 80.64 & 81.75 & 80.64 & 100 & 82.01 & 81.02 & 82.01 & 81.02 & 100 & 81.40 & 80.47 & 81.40 & 80.47 \\
Uncertainty & 100 & 83.48 & 81.13 & 83.48 & 81.13 & 100 & 84.39 & 82.65 & 84.39 & 82.65 & 100 & 84.26 & 82.13 & 84.26 & 82.13 \\
1-shot w/ Def & 100 & \colorbox{blue!30}{84.34} & \colorbox{blue!30}{82.79} & \colorbox{blue!30}{84.34} & \colorbox{blue!30}{82.79} & 100 & 83.95 & 82.44 & 83.95 & 82.44 & 100 & 83.65 & 82.11 & 83.65 & 82.11 \\
\midrule
\multirow{3}{*}{\textbf{Method}} & \multicolumn{5}{c|}{\textbf{5\_shot}} & \multicolumn{5}{c|}{\textbf{category\_definition}} & \multicolumn{5}{c}{\textbf{numerical}} \\
\cmidrule(lr){2-6} \cmidrule(lr){7-11} \cmidrule(lr){12-16}
 & fmt-suc & fmt-suc & fmt-suc & overall & overall & fmt-suc & fmt-suc & fmt-suc & overall & overall & fmt-suc & fmt-suc & fmt-suc & overall & overall \\
 & ratio & acc & macro-f1 & acc & macro-f1 & ratio & acc & macro-f1 & acc & macro-f1 & ratio & acc & macro-f1 & acc & macro-f1 \\
\midrule
Zero-shot & 100 & \colorbox{blue!30}{84.04} & \colorbox{blue!30}{82.93} & \colorbox{blue!30}{84.04} & \colorbox{blue!30}{82.93} & 100 & 82.22 & 81.44 & 82.22 & 81.44 & 0 & -- & -- & -- & -- \\
1-shot & 100 & 83.61 & 82.85 & 83.61 & 82.85 & 100 & 83.09 & 81.99 & 83.09 & 81.99 & 0 & -- & -- & -- & -- \\
3-shot & 100 & 84.95 & \colorbox{blue!30}{\textbf{84.01}} & 84.95 & \colorbox{blue!30}{\textbf{84.01}} & 100 & 81.79 & 20.30 & 81.79 & 20.30 & 0 & -- & -- & -- & -- \\
5-shot & 100 & 84.04 & 83.32 & 84.04 & 83.32 & 100 & 83.87 & 82.55 & 83.87 & 82.55 & 2.51 & 82.76 & 31.31 & 2.08 & 0.79 \\
Definition & 100 & 83.09 & 82.19 & 83.09 & 82.19 & 100 & \colorbox{blue!30}{83.26} & \colorbox{blue!30}{82.30} & \colorbox{blue!30}{83.26} & \colorbox{blue!30}{82.30} & 26.38 & 82.13 & 34.14 & 21.67 & 9.01 \\
Numerical & 100 & 54.67 & 10.89 & 54.67 & 10.89 & 100 & 12.41 & 8.79 & 12.41 & 8.79 & 100 & \colorbox{blue!30}{83.17} & \colorbox{blue!30}{81.89} & \colorbox{blue!30}{83.17} & \colorbox{blue!30}{81.89} \\
Similar-3-shot & 100 & 81.75 & 78.96 & 81.75 & 78.96 & 100 & 80.19 & 77.20 & 80.19 & 77.20 & 19.20 & 76.80 & 48.53 & 14.75 & 9.32 \\
Fixed-3-shot & 100 & 82.53 & 81.73 & 82.53 & 81.73 & 100 & 81.92 & 81.00 & 81.92 & 81.00 & 1.04 & 79.17 & 29.46 & 0.82 & 0.31 \\
Uncertainty & 100 & 84.60 & 82.77 & 84.60 & 82.77 & 100 & 84.47 & 83.09 & 84.47 & 83.09 & 0 & -- & -- & -- & -- \\
1-shot w/ Def & 100 & 84.17 & 82.47 & 84.17 & 82.47 & 100 & 84.04 & 83.16 & 84.04 & 83.16 & 0.26 & 83.33 & 45.45 & 0.22 & 0.12 \\
\midrule
\multirow{3}{*}{\textbf{Method}} & \multicolumn{5}{c|}{\textbf{similar\_3\_shot}} & \multicolumn{5}{c|}{\textbf{zero\_shot}} & \multicolumn{5}{c}{\textbf{ppl}} \\
\cmidrule(lr){2-6} \cmidrule(lr){7-11} \cmidrule(lr){12-16}
 & fmt-suc & fmt-suc & fmt-suc & overall & overall & fmt-suc & fmt-suc & fmt-suc & overall & overall & fmt-suc & fmt-suc & fmt-suc & overall & overall \\
 & ratio & acc & macro-f1 & acc & macro-f1 & ratio & acc & macro-f1 & acc & macro-f1 & ratio & acc & macro-f1 & acc & macro-f1 \\
\midrule
Zero-shot & 100 & 83.09 & 82.00 & 83.09 & 82.00 & 100 & 82.74 & 81.73 & 82.74 & 81.73 & 100 & 63.58 & 40.28 & 63.58 & 40.28 \\
1-shot & 100 & 83.00 & 81.80 & 83.00 & 81.80 & 100 & 83.04 & 81.95 & 83.04 & 81.95 & 100 & 71.76 & 52.92 & 71.76 & 52.92 \\
3-shot & 100 & 84.04 & 82.52 & 84.04 & 82.52 & 100 & 83.09 & 80.90 & 83.09 & 80.90 & 100 & 70.85 & 48.13 & 70.85 & 48.13 \\
5-shot & 100 & 83.87 & 83.34 & 83.87 & 83.34 & 100 & 83.61 & 82.62 & 83.61 & 82.62 & 100 & 66.35 & 43.04 & 66.35 & 43.04 \\
Definition & 100 & 82.66 & 81.43 & 82.66 & 81.43 & 100 & 82.79 & 81.71 & 82.79 & 81.71 & 100 & 66.96 & 52.74 & 66.96 & 52.74 \\
Numerical & 100 & 57.18 & 8.77 & 57.18 & 8.77 & 100 & 10.68 & 4.02 & 10.68 & 4.02 & 100 & 75.30 & 70.38 & 75.30 & 70.38 \\
Similar-3-shot & 100 & 81.57 & \colorbox{blue!30}{79.68} & 81.57 & \colorbox{blue!30}{79.68} & 100 & 81.49 & 77.98 & 81.49 & 77.98 & 100 & 65.27 & 41.84 & 65.27 & 41.84 \\
Fixed-3-shot & 100 & \colorbox{blue!30}{83.17} & \colorbox{blue!30}{82.49} & \colorbox{blue!30}{83.17} & \colorbox{blue!30}{82.49} & 100 & 82.70 & 81.56 & 82.70 & 81.56 & 100 & 67.56 & 48.30 & 67.56 & 48.30 \\
Uncertainty & 100 & 83.87 & 81.84 & 83.87 & 81.84 & 100 & \colorbox{blue!30}{\textbf{85.08}} & \colorbox{blue!30}{83.74} & \colorbox{blue!30}{\textbf{85.08}} & \colorbox{blue!30}{83.74} & 100 & 72.97 & 62.39 & 72.97 & 62.39 \\
1-shot w/ Def & 100 & 83.61 & 81.45 & 83.61 & 81.45 & 100 & 83.78 & 82.84 & 83.78 & 82.84 & 100 & 70.59 & 51.07 & 70.59 & 51.07 \\
\bottomrule
\end{tabular}
\end{table*}

\section{Experiments}
\subsection{Experimental Setup}
\paragraph{Datasets}
We conducted comprehensive experiments on seven datasets, including four public benchmarks (EC, EIC, IFLYTEK, and TNEWS) and three proprietary datasets (Query Intent, Search Correlation, and Query Taxonomy). EC focuses on sentiment detection, whereas EIC classifies the type of edits between sentence pairs, a task on which generative classifiers have previously performed poorly \citep{ruan2024large}. IFLYTEK assigns app descriptions to as many as 120 categories, and TNEWS categorizes news headlines by topic; both are widely used multi‑class benchmarks. Our proprietary datasets further extend the evaluation: Query Intent (QI) predicts user intent at both coarse and fine granularities across roughly 30 labels, Search Correlation (SC) evaluates the relevance between a query and a text passage, and Query Taxonomy (QT) performs multi‑label semantic tagging, since a single query may map to multiple categories. More detailed descriptions of all datasets are provided in Appendix~\ref{Data Statistics}.
\paragraph{Metrics}
The performance of our models is evaluated using accuracy (Acc.) and macro-F1. Accuracy measures the ratio of correct predictions to total predictions, while macro-F1 is the average of per-class F1-scores, assigning equal weight to each class. For each inference strategy, we report five metrics: \textit{fmt-suc ratio} (percentage of format-matched outputs), \textit{fmt-suc accuracy} and \textit{fmt-suc macro-F1} (computed only on format-matched outputs), and \textit{overall accuracy} and \textit{overall macro-F1} (calculated over all predictions). The \textbf{overall accuracy} and \textbf{overall macro-F1} serve as the primary indicators of task performance. When \textit{fmt-suc ratio} is less than 100\%, format-success metrics are highlighted only if their corresponding overall metrics also achieve best performance.
\paragraph{Parameter Setting}
We used Qwen-2.5-7B-Instruct \citep{yang2024qwen2} as our base language model. This open-source model achieves non-trivial performance on the classification task while still leaving room for improvement, making it an ideal testbed for our study. We constructed the training dataset using the prompt strategy described in Section~\ref{3.1} and tested each trained model across all these prompt types, yielding approximately $10 \times 10$ total combinations. For RL, we used Reinforce++ and its training framework OpenRLHF \citep{hu2025reinforce++}, in our experiments.
\paragraph{Baselines}
Since our approach employs generative classification, we adopt the traditional discriminative method using value head on public datasets as a robust baseline. Specifically, we utilize the results reported by \cite{ruan2024large} for the EC and EIC datasets, and by \cite{xu2020clue} for the IFLYTEK and TNEWS datasets. Additionally, to illustrate that combining different prompt strategies during training and inference can yield superior performance, we introduce an additional commonaly used naive SFT baseline, using a zero-shot prompt strategy for both stages (i.e. training and evaluating the model exclusively with the zero-shot prompt). 

\subsection{Main Results}
We adopt a generative classification paradigm based on a large language model. The base model is fine-tuned using various prompt strategies and evaluated under each strategy at inference time. Since a single dataset can yield nearly one hundred training–inference combinations, Table~\ref{tab:results} reports only the best result for each training strategy to enable comprehensive analysis. For illustration, we present the full set of combinations for the EIC dataset in Table~\ref{tab:EIC}, and results for the remaining datasets are provided in Appendix~\ref{sft results}.

As shown in Table \ref{tab:results}, GenCLS++ surpasses every discriminative baseline on the public datasets, underscoring the strength of generative approaches to classification. Moreover, for most training prompts, \textbf{switching to an alternative inference prompt yields additional gains in both accuracy and macro-F1}. Figure \ref{fig:acc_improve} visualizes these improvements, demonstrating that—\textbf{regardless of the strategy employed during training—experimenting with a different inference strategy typically leads to superior performance}. Moreover, applying RL to a model that has already undergone SFT yields additional gains. Although the naive SFT baselines for EC and Query Intent are already strong, exceeding 90\% accuracy, GenCLS++ still achieves an average relative accuracy improvement of 3.46\% on all seven datasets and 4.00\% on the four public datasets. Notably, GenCLS++ delivers a 6.10\% relative accuracy improvement on the IFLYTEK dataset, underscoring its effectiveness.

Further analysis reveals a consistent pattern: adding labeled examples to the training prompt (few-shot learning) consistently outperforms training without examples, and this advantage holds in both zero-shot and few-shot evaluations. Additionally, randomly sampled examples yield higher scores than a fixed set of examples.

Although the optimal inference prompt is not always identical to the training prompt, two clear tendencies emerge: 1) If the training prompt includes few-shot examples, the highest scores are achieved when the inference prompt also provides examples. 2) If the training prompt omits examples, a zero-shot inference prompt is usually the stronger choice.

These findings underscore that prompt design should be considered jointly for training and inference, rather than in isolation. Furthermore, using reinforcement learning to further enhance the performance of generative models on classification tasks is a promising approach, which we will analyze in more detail in Section~\ref{5}.

\section{Analysis}
\label{5}
\subsection{Effectiveness of the Policy Warm-up}
\label{5.1}
To equip the policy model with fundamental classification capabilities, we first apply fine-tuning on the training data, which we refer to as the "warm-up" phase. We then conduct an ablation study using Qwen-2.5-7B-Instruct on several public benchmarks to evaluate the impact of this phase. As reported in Table~\ref{tab:warm-up}, incorporating a warm-up phase provides a significant performance boost in subsequent RL training, with an average relative accuracy improvement of 18.18\% over initializing RL directly from the base model. This indicates that allowing the policy model to acquire essential classification skills through supervised fine-tuning (SFT) before RL effectively raises the ceiling for achievable performance, demonstrating the importance of pre-training in enhancing RL outcomes.

\begin{table}[ht]
    \centering
    \small
    \caption{Experimental Results (Accuracy \& macro-F1 Score, \%). \textbf{Bold} indicates the best result.}
    \label{tab:warm-up}
    \begin{tabular}{lcccccccc}
    \toprule
    & \multicolumn{2}{c}{EC} & \multicolumn{2}{c}{EIC} & \multicolumn{2}{c}{IFLYTEK} & \multicolumn{2}{c}{TNEWS} \\
    \cmidrule(lr){2-3}\cmidrule(lr){4-5}\cmidrule(lr){6-7}\cmidrule(lr){8-9}
    \textbf{Training Method}  &Acc. &macro-F1  &Acc. &macro-F1  &Acc. &macro-F1 &Acc. &macro-F1 \\
    \midrule
    Base model      &68.75 &33.94  &55.02 &49.85  &57.41 &25.93  &58.13 &22.92 \\
    \quad + RL      &76.90 &69.52  &62.28 &47.87  &59.79 &29.88  &61.92 &59.56 \\
    \quad + warm up &94.15 &89.83  &85.08 &83.74  &63.76 &47.85  &63.30 &61.31 \\
    \quad\quad + RL &\bf{94.50} &\bf{90.57}  &\bf{85.86} &\bf{84.72}  &\bf{64.91} &\bf{49.27}  &\bf{64.04} &\bf{62.35} \\
    \bottomrule
    \end{tabular}
\end{table}

\subsection{Does the reasoning process really help classification?}
\label{5.2}
In this subsection, we present further discussions on the reasoning process in fine-tuning for classification. We explored the following settings: <Class$\rightarrow$Reason>, <Reason$\rightarrow$Class>, and <Think$\rightarrow$Reason$\rightarrow$Class>, which define the order in which the model outputs its responses. For example, in the <Reason$\rightarrow$Class> setting, the model first explains its reasoning and then predicts the classification result. Here, “Think” represents a longer, more elaborate thought, while “Reason” represents a more concise explanation. 

We chose the EIC dataset for our research because identifying the categories of editing intent requires comparing the changes before and after sentences, which requires the model to use reasoning ability more than other classification tasks. We used DeepSeek-R1 \citep{guo2025deepseek} to generate reasoning for the training set in the <Think$\rightarrow$Reason$\rightarrow$Class> format and then manually converted these outputs into the three settings mentioned above. For each question, we sampled three times; the reasoning process was considered valid only if all three outputs were correct. We then performed supervised fine-tuning (SFT) of the base model on these three datasets, and the results are shown in Table~\ref{tab:results_reasoning}. Interestingly, contrary to intuition, having the model provide its classification result first led to higher accuracy compared to the other two approaches.

To delve deeper into the reasoning process for classification, we ran RL from several starting checkpoints: the base model, a SFT model without reasoning (No-CoT SFT), an SFT model with reasoning (CoT SFT), and a two-stage model obtained by further fine-tuning the No-CoT SFT model on CoT data. The results indicate the following: 1) Regardless of the starting model, RL consistently enhances classification performance. 2) Before applying CoT data for SFT, conducting an initial SFT stage using No-CoT data leads to better performance improvements following RL. 3) Interestingly, the best performance is attained by directly applying RL to the No-CoT SFT model and allowing the model to predict the answer without performing reasoning beforehand.

\begin{figure*}[!ht] 
\centering
\includegraphics[width=\linewidth]{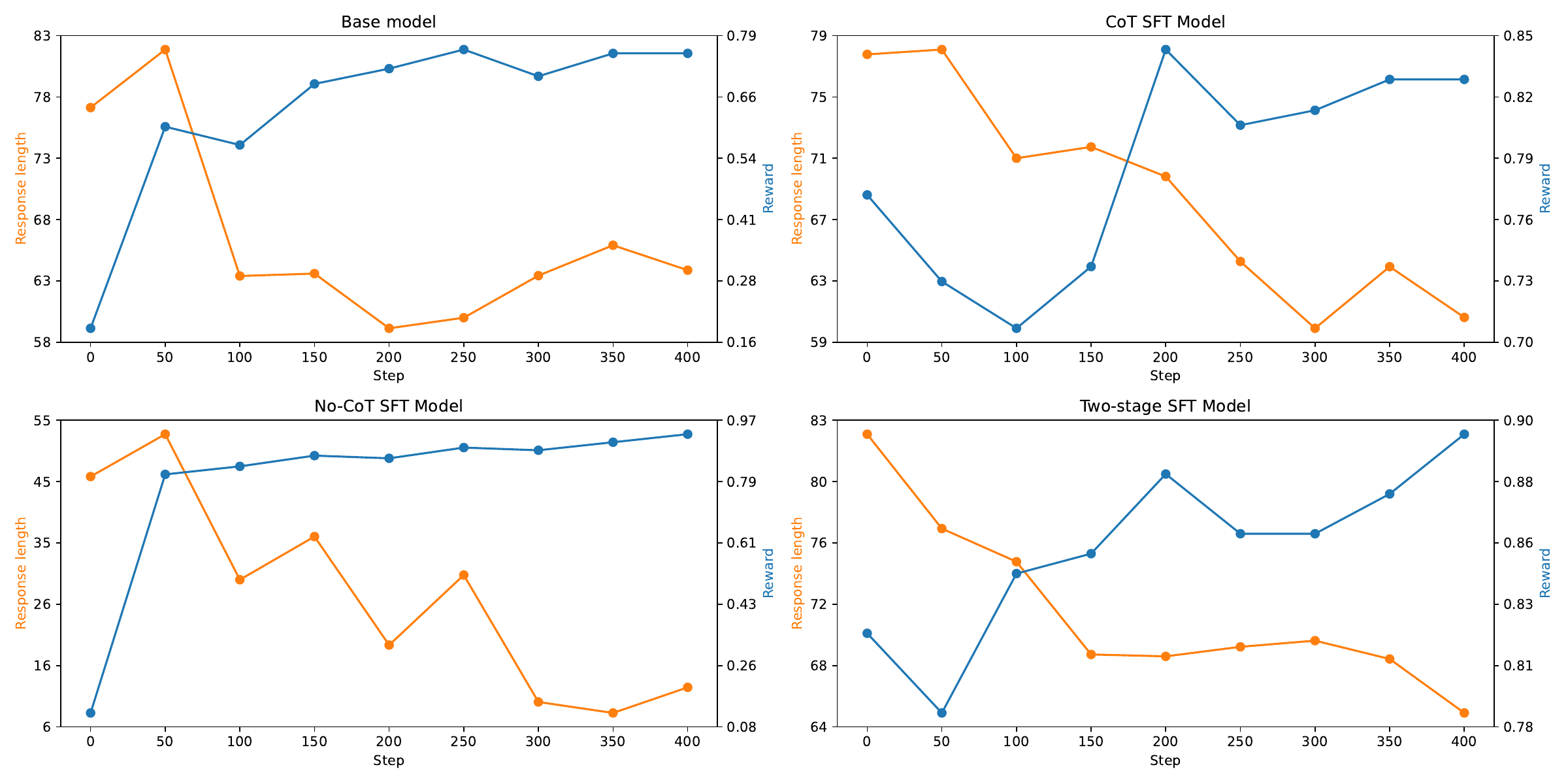}
\caption{Comparison of Different Models: Response Length vs. Reward over Steps}
\label{fig:reward_response_length}
\end{figure*}

We subsequently examined how the model’s output length and behavior evolved during RL training. As shown in Figure \ref{fig:reward_response_length}, the length of the generated “reason” gradually decreased, indicating that the model pruned away unnecessary reasoning steps. These results suggest that the model progressively learns to simplify its reasoning and that extensive deliberation is not always beneficial for producing correct answers. In classification tasks in particular, the RL signal appears to teach the model that directly producing the answer is sufficient.

\begin{table}[ht]
    \centering
    \small
    \caption{Experimental Results on the EIC dataset (Accuracy \& macro-F1 Score, \%). \colorbox{blue!30}{Blue} indicates the best inference strategy for current training method. \textbf{Bold} indicates the best result.}
    \label{tab:results_reasoning}
    \begin{tabular}{lcccccc}
    \toprule
    & \multicolumn{2}{c}{Class$\rightarrow$Reason} & \multicolumn{2}{c}{Reason$\rightarrow$Class} & \multicolumn{2}{c}{Think$\rightarrow$Reason$\rightarrow$Class} \\
    \cmidrule(lr){2-3}\cmidrule(lr){4-5}\cmidrule(lr){6-7}
    \textbf{Training Method} & Acc. & macro-F1 & Acc. & macro-F1 & Acc. & macro-F1 \\
    \midrule
    Base model                             &17.95 &13.86  &\cellcolor{blue!30}30.54 &\cellcolor{blue!30}5.69  &29.46 &7.27 \\
    \quad + RL                            &53.33 &42.46  &\cellcolor{blue!30}62.28 &\cellcolor{blue!30}47.87  &46.58 &19.60 \\
    \midrule
    SFT (Class$\rightarrow$Reason)                     &\cellcolor{blue!30}76.73 &\cellcolor{blue!30}76.12  &57.05 &51.46  &57.40 &52.34 \\
    SFT (Think$\rightarrow$Reason$\rightarrow$Class)               &37.98 &36.15  &53.81 &47.14  &\cellcolor{blue!30}57.35 &\cellcolor{blue!30}50.45 \\
    \midrule
    SFT (Reason$\rightarrow$Class)                     &64.62 &64.67  &\cellcolor{blue!30}71.37 &\cellcolor{blue!30}66.26  &70.11 &45.60 \\
    \quad + RL                            &77.12 &76.00  &\cellcolor{blue!30}79.50 &\cellcolor{blue!30}78.06  &78.63 &76.00 \\
    \midrule
    SFT (No-CoT)                           &\cellcolor{blue!30}76.60 &\cellcolor{blue!30}75.64  &69.81 &62.03 &74.57 &33.76 \\
    \quad + RL                            &\cellcolor{blue!30}\bf{85.86} &\cellcolor{blue!30}\bf{84.72}  &85.64 &84.52 &84.60 &83.26 \\
    \quad + SFT (Reason$\rightarrow$Class)            &63.62 &52.13  &\cellcolor{blue!30}76.08 &\cellcolor{blue!30}73.63 &-     &-     \\
    \quad\quad + RL                      &0.43  &0.43   &\cellcolor{blue!30}84.04 &\cellcolor{blue!30}82.67 &33.48 &31.90 \\
    \bottomrule
    \end{tabular}
\end{table}

\begin{figure*}[!ht] 
    \centering
    \includegraphics[width=\linewidth]{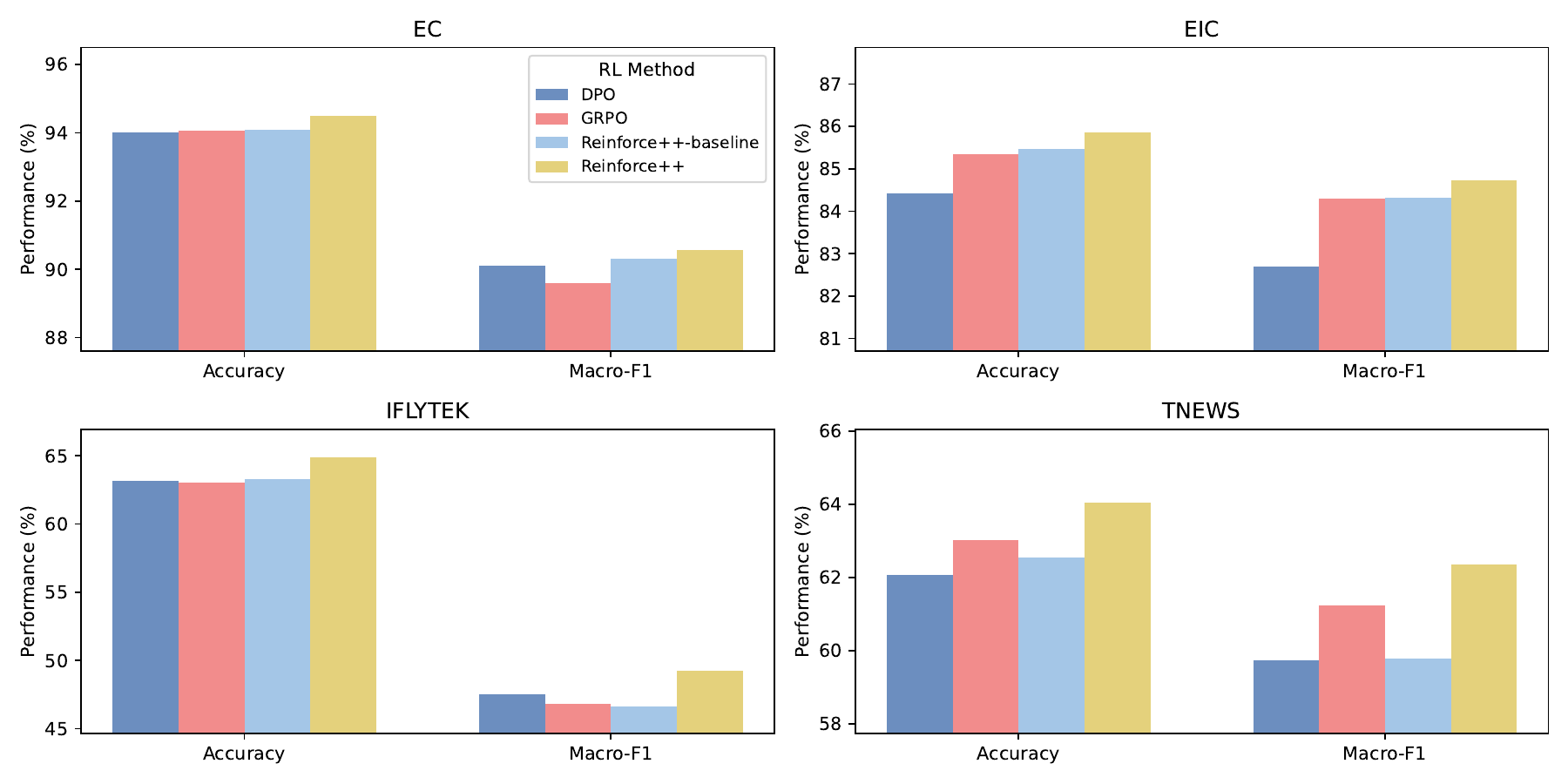}
    \caption{Comparison of model performance across different RL algorithms.}
    \label{fig:diff_rl}
\end{figure*}

\subsection{Different RL Algorithms}
\label{5.3}
We further explored the impact of different RL algorithms on model performance by analyzing GRPO \citep{shao2024deepseekmath}, Reinforce++-baseline \citep{hu2025reinforce++}, and Reinforce++ \citep{hu2025reinforce++}, with DPO \citep{rafailov2023direct} included as an off-policy comparator. As shown in Table \ref{fig:diff_rl}, all on-policy methods outperform DPO. Notably, Reinforce++ yields the largest gains and, unlike GRPO, requires no batch sampling of candidate responses during training—making it the most efficient choice.

\section{Conclusion}
In this paper, we investigate the use of LLMs as generative classifiers. By systematically exploring a variety of prompt strategies during both training and inference, coupled with the integration of reinforcement learning, we enhance the intrinsic generative capabilities of LLMs for classification tasks. GenCLS++ achieves an average relative accuracy improvement of +3.46\% across seven benchmark datasets compared to the naive SFT baseline. Notably, our experiments show that while explicit reasoning steps enhance performance on complex tasks, they do not yield significant benefits in classification settings. In future work, we aim to evaluate whether these findings generalize to models of varying scales and to explore novel techniques that can further push the performance limits of generative classifiers.

\clearpage
\bibliographystyle{apalike}
\bibliography{references}

\clearpage
\appendix
\section{Prompt Example}
\label{A}
To illustrate, we use the EC dataset to showcase the prompt strategies outlined in Section~\ref{3.1}.
\paragraph{Zero-shot \& Few-shot} We adopt the 3-shot setting as a representative example for both the zero-shot and few-shot series.
\begin{tcolorbox}[boxrule=0.5mm, left=1mm, right=1mm, top=1mm, bottom=1mm]
You are a professional sentiment classification expert. There is now a piece of text that requires your sentiment classification.

Optional categories: [sadness, joy, love, anger, fear, surprise]

Format requirement: Please output in the format \verb|Category: xxx| (where \texttt{xxx} is the corresponding category label).

\medskip
\textbf{Example 1:}\\
Text: i feel now i am not giving all of me to christ and i want to be devoted\\
Category: love

\textbf{Example 2:}\\
Text: i find myself feeling shocked hearing that word spoken out loud in my own lounge room\\
Category: surprise

\textbf{Example 3:}\\
Text: i feel pathetic and that i shouldn’t make myself feel this way\\
Category: sadness

\medskip
\textbf{Current case:}\\
Text: i feel like a low life mooching off everyone

Please output the category for the text according to the format requirement.
\end{tcolorbox}

\paragraph{Definition} We prepend concise text definitions of each target category to the prompt and further explore an alternative strategy by adding an extra example.
\begin{tcolorbox}[boxrule=0.5mm, left=1mm, right=1mm, top=1mm, bottom=1mm]
You are a professional sentiment classification expert. There is now a piece of text that requires your sentiment classification.

Optional categories: [sadness, joy, love, anger, fear, surprise]  

Format requirement: Please output in the format \verb|Category: xxx| (where \texttt{xxx} is the corresponding category label).

\medskip
Sentiment category definitions:\\
\textbf{sadness}: expresses loss, sorrow, frustration, etc., involving farewells, failures, regrets, etc.\\
\textbf{joy}: conveys happiness, cheerfulness, satisfaction, etc., including celebrations, success, and pleasure from good things.\\
\textbf{love}: reflects romantic love, familial love, friendship, etc., involving care, admiration, attachment, etc.\\
\textbf{anger}: contains strong negative feelings like anger, annoyance, indignation, involving injustice, conflict, frustration, etc.\\
\textbf{fear}: shows fear, worry, anxiety, etc., involving danger, uncertainty, psychological pressure, etc.\\
\textbf{surprise}: expresses the unexpected, astonishment, amazement, etc., including sudden events or information beyond expectations.

\medskip
{\color{gray}%
\textbf{Example 1:}\\
Text: i walk in the door to my house i feel happy\\
Category: joy
}

\medskip
\textbf{Current case:}\\
Text: i was feeling like a beluga whale and quite grouchy

Please output the category for the text according to the format requirement.
\end{tcolorbox}

\clearpage
\paragraph{Numerical} For this type, we assign a numerical label to each category and instruct the model to output the corresponding number.
\begin{tcolorbox}[boxrule=0.5mm, left=1mm, right=1mm, top=1mm, bottom=1mm]
You are a professional sentiment classification expert. There is now a piece of text that requires your sentiment classification.

Optional categories: sadness: 0, joy: 1, love: 2, anger: 3, fear: 4, surprise: 5

Format requirement: Please output in the format \verb|Category: xxx| (where \texttt{xxx} is the corresponding numeric label).

\medskip
\textbf{Current case:}\\
Text: i reply i do my best to reply to questions but feel free to contact me via twitter isobelmeg xx

Please output the category for the text according to the format requirement.
\end{tcolorbox}

\paragraph{Uncertainty} We introduce a new class, “Uncertain,” and employ a fine-tuned model to label training examples that cannot be classified with high confidence. This prompt strategy is used exclusively during dataset construction; during inference, the model is restricted to predicting only the original classes.
\begin{tcolorbox}[boxrule=0.5mm, left=1mm, right=1mm, top=1mm, bottom=1mm]
You are a professional sentiment classification expert. There is now a piece of text that requires your sentiment classification.

Optional categories: [sadness, joy, love, anger, fear, surprise]

Format requirement: Please output in the format \verb|Category: xxx| (where \texttt{xxx} is the corresponding category label; if unsure, please reply \verb|"Category: uncertain"|).

\medskip
\textbf{Current case:}\\
Text: i forgive myself that i have accepted and allowed myself to forget that i decide and thus i was decided to feel groggy this morning

Please output the category for the text according to the format requirement.
\end{tcolorbox}

\section{Data Statistics}
\label{Data Statistics}
The detailed statistics for all datasets are shown in Table~\ref{tab:prompt_data} and ~\ref{tab:data_sta}

\begin{table*}
\caption{Average token length statistics across different tasks and prompt types. Both prompt tokens and response tokens represent average values per example.}
\label{tab:prompt_data}
\centering
\scriptsize
\setlength{\tabcolsep}{2.8pt}
\begin{tabular}{@{}l|cc|cc|cc|cc|cc@{}}
\toprule
\multirow{2}{*}{\textbf{Dataset}} & \multicolumn{2}{c|}{\textbf{Zero-shot}} & \multicolumn{2}{c|}{\textbf{1-shot}} & \multicolumn{2}{c|}{\textbf{3-shot}} & \multicolumn{2}{c|}{\textbf{5-shot}} & \multicolumn{2}{c}{\textbf{Definition}} \\
\cmidrule(lr){2-3} \cmidrule(lr){4-5} \cmidrule(lr){6-7} \cmidrule(lr){8-9} \cmidrule(lr){10-11}
 & Prompt & Response & Prompt & Response & Prompt & Response & Prompt & Response & Prompt & Response \\
 & tokens & tokens & tokens & tokens & tokens & tokens & tokens & tokens & tokens & tokens \\
\midrule
Query Intent & 282.4 & 6.0 & 305.9 & 6.0 & 344.8 & 6.0 & 383.8 & 6.0 & 719.4 & 6.0 \\
Query Taxonomy & 1225.4 & 7.7 & 1250.5 & 7.7 & 1292.6 & 7.7 & 1334.8 & 7.7 & - & - \\
Search Correlation & 458.6 & 5.2 & 843.0 & 5.2 & 1605.9 & 5.2 & 2367.5 & 5.2 & 1261.6 & 5.2 \\
EIC & 141.6 & 4.1 & 222.1 & 4.1 & 375.9 & 4.1 & 523.6 & 4.1 & 287.6 & 4.1 \\
EC & 86.0 & 3.5 & 121.4 & 3.5 & 184.4 & 3.5 & 246.9 & 3.5 & 229.0 & 3.5 \\
TNEWS & 119.8 & 4.7 & 152.3 & 4.7 & 209.3 & 4.7 & 266.4 & 4.7 & 631.8 & 4.7 \\
IFLYTEK & 786.9 & 3.6 & 984.3 & 3.6 & 1370.2 & 3.6 & 1761.1 & 3.6 & 3481.9 & 3.6 \\
\midrule
\multirow{2}{*}{\textbf{Dataset}} & \multicolumn{2}{c|}{\textbf{Numerical}} & \multicolumn{2}{c|}{\textbf{Similar-3-shot}} & \multicolumn{2}{c|}{\textbf{Fixed-3-shot}} & \multicolumn{2}{c|}{\textbf{Uncertainty}} & \multicolumn{2}{c}{\textbf{1-shot w/ Def}} \\
\cmidrule(lr){2-3} \cmidrule(lr){4-5} \cmidrule(lr){6-7} \cmidrule(lr){8-9} \cmidrule(lr){10-11}
 & Prompt & Response & Prompt & Response & Prompt & Response & Prompt & Response & Prompt & Response \\
 & tokens & tokens & tokens & tokens & tokens & tokens & tokens & tokens & tokens & tokens \\
\midrule
Query Intent & 446.4 & 3.7 & 344.2 & 6.0 & 342.4 & 6.0 & 291.4 & 6.0 & 741.9 & 6.0 \\
Query Taxonomy & 2746.4 & 6.9 & 1292.0 & 7.7 & 1294.4 & 7.7 & 1225.4 & 6.6 & - & - \\
Search Correlation & 475.6 & 4.0 & 1921.0 & 5.2 & 1740.6 & 5.2 & 473.1 & 5.1 & 1653.0 & 5.2 \\
EIC & 155.6 & 3.0 & 400.6 & 4.1 & 596.5 & 4.1 & 145.6 & 4.1 & 368.1 & 4.1 \\
EC & 103.0 & 3.0 & 163.4 & 3.5 & 186.0 & 3.5 & 95.0 & 3.4 & 264.4 & 3.5 \\
TNEWS & 193.8 & 5.0 & 210.7 & 4.7 & 194.8 & 4.7 & 128.8 & 4.5 & 665.3 & 4.7 \\
IFLYTEK & 1272.9 & 4.0 & 1476.9 & 3.6 & 1215.9 & 3.6 & 795.9 & 3.5 & 3682.3 & 3.6 \\
\bottomrule
\end{tabular}
\label{tab:token_sta}
\end{table*}

\begin{table}
  \centering
  \caption{Dataset Statistics}
  \label{tab:data_sta}
  \begin{tabular}{lcccc}
    \hline
    \textbf{Dataset} & \textbf{Train samples} & \textbf{Test samples} & \textbf{Class count} & \textbf{Classification type} \\
    \hline
    Query Intent & 100,000 & 10,000 & 33 & Single-label \\
    Query Taxonomy & 200,000 & 10,000 & 326 & Multi-label \\
    Search Correlation & 311,446 & 2,997 & 5 & Single-label \\
    EIC & 7,478 & 2,312 & 5 & Single-label \\
    EC & 16,000 & 2,000 & 6 & Single-label \\
    TNEWS & 53,360 & 10,000 & 15 & Single-label \\
    IFLYTEK & 12,133 & 2,599 & 119 & Single-label \\
    \hline
  \end{tabular}
\end{table}

\section{SFT Results}
\label{sft results}
\subsection{Performance Gains from Prompt Strategy Switching}
\begin{figure*}[!ht] 
    \centering
    \includegraphics[width=\linewidth]{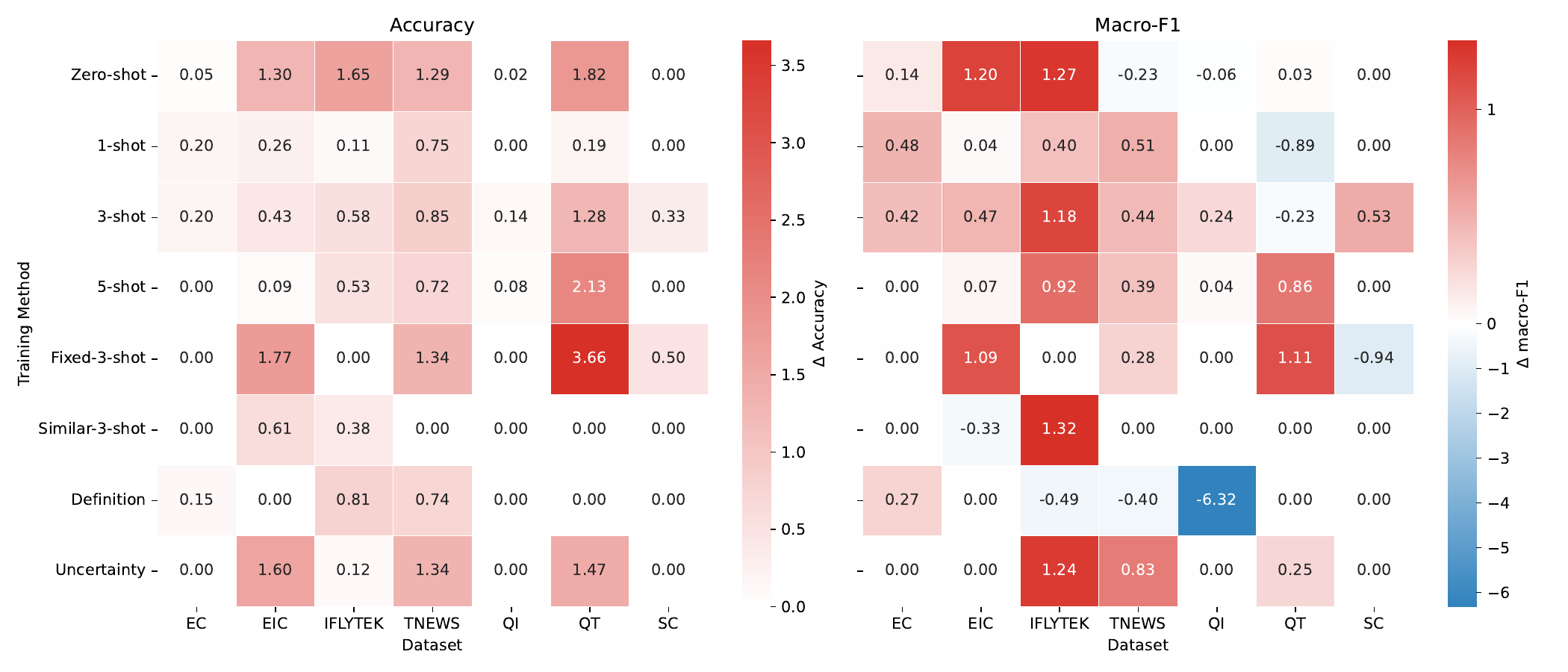}
    \caption{Visualization of the improvement achieved by changing the inference prompt strategy. Left: improvement in accuracy. Right: improvement in macro-F1. The x-axis represents the dataset, and the y-axis represents the different training strategies. Improvements are highlighted in red, while decreases are shown in blue.}
    \label{fig:acc_improve}
\end{figure*}

We visualize the improvement brought by using the best-performing inference prompt strategy compared to reusing the same strategy as in training in Figure~\ref{fig:acc_improve}. As shown, our approach consistently improves both accuracy and macro-F1 across nearly all tasks and training prompt strategies.

\subsection{Effect of Retrieval Relevance in Few-Shot Inference}
\begin{figure*}[!ht] 
    \centering
    \includegraphics[width=\linewidth]{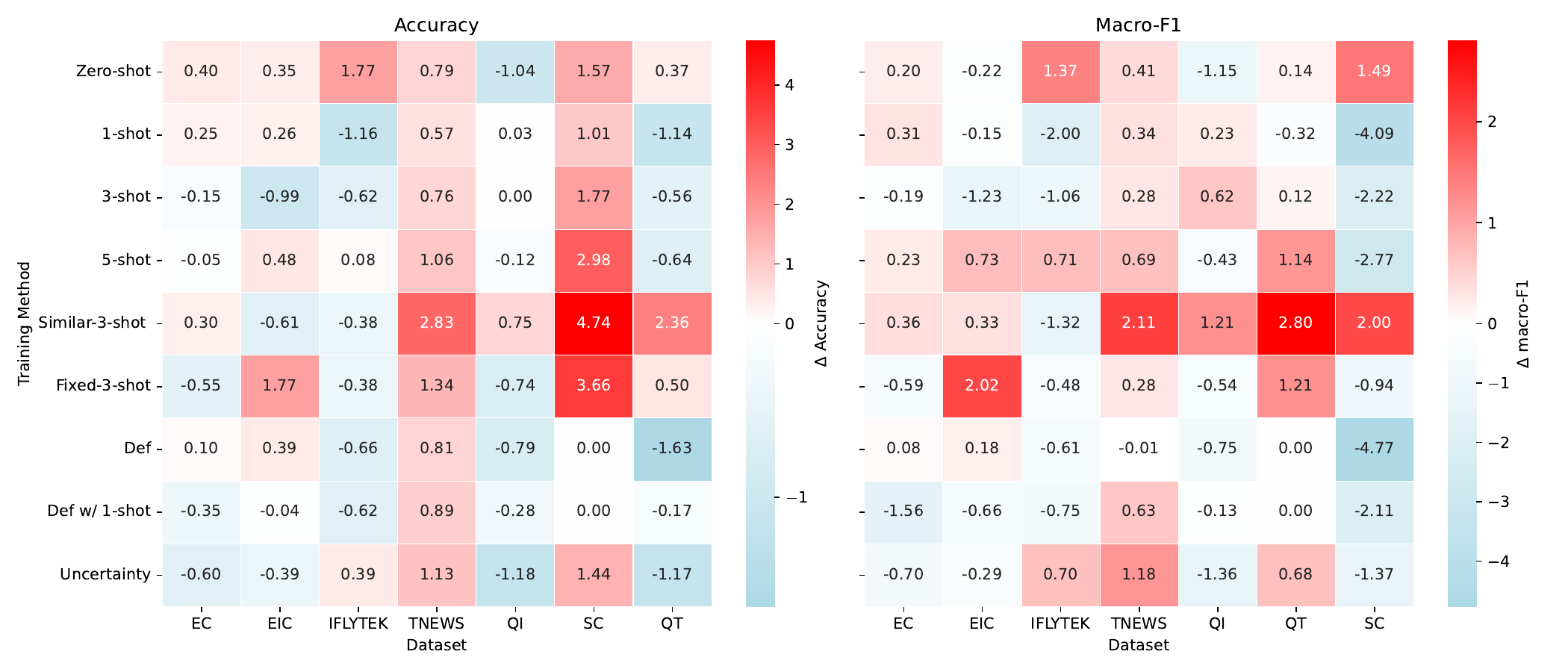}
    \caption{Visualization of the improvement achieved by changing the inference prompt strategy from fixed-3-shot to similar-3-shot. Left: improvement in accuracy. Right: improvement in macro-F1. The x-axis represents the dataset, and the y-axis represents the different training strategies. Improvements are highlighted in red, while decreases are shown in blue.}
    \label{fig:fix-similar}
\end{figure*}

We further compared the impact of two different inference strategies on model performance. As shown in Figure~\ref{fig:fix-similar}, using similar few-shot examples does not always lead to better results across all tasks. We found that when the retrieval strategy is related to the category, as is the case with the TNEWS dataset, the retrieved examples can enhance performance. However, for tasks involving relationships between multiple texts, where the retrieval strategy is unrelated to classification, including seemingly similar examples may actually degrade performance.

\subsection{Perplexity-Based Strategy Evaluation}
\begin{figure*}[!ht] 
    \centering
    \includegraphics[width=\linewidth]{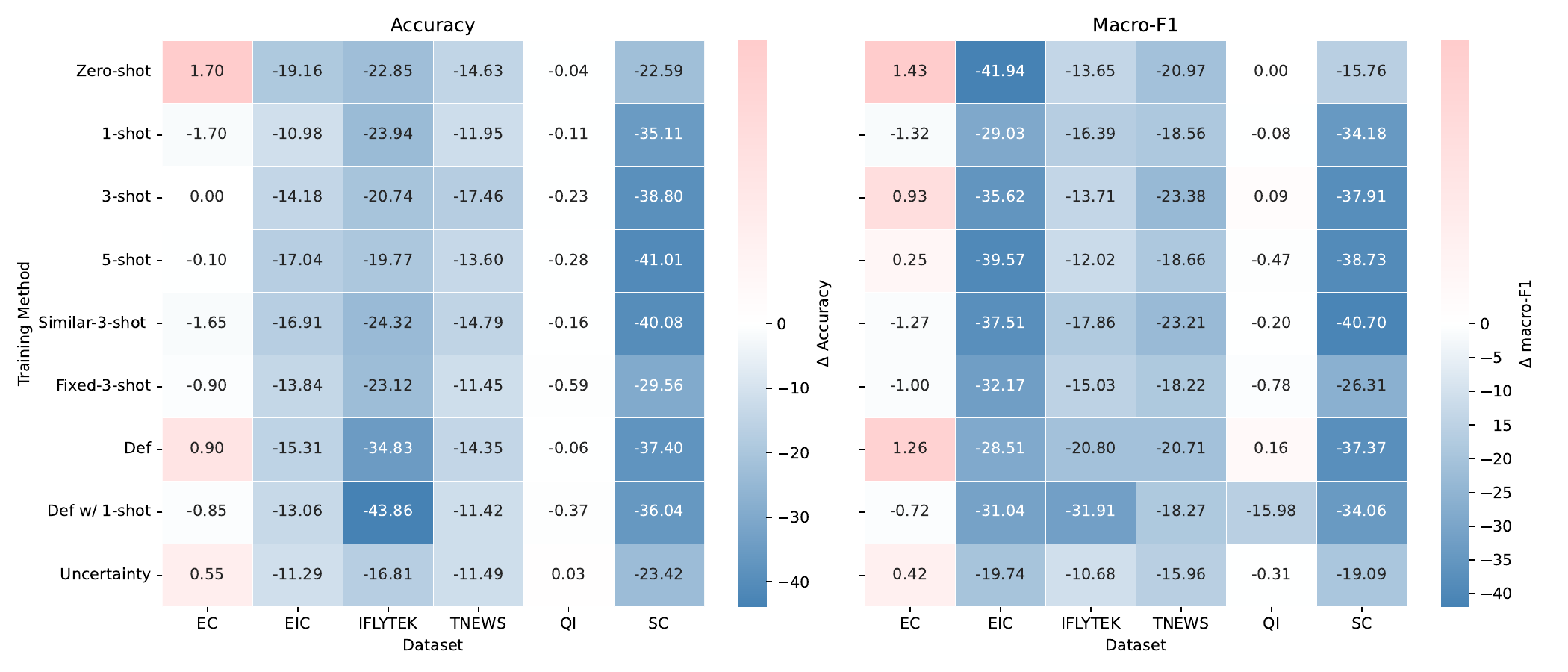}
    \caption{Visualization of the improvement achieved by changing the inference prompt strategy from fixed-3-shot to perplexity. Left: improvement in accuracy. Right: improvement in macro-F1. The x-axis represents the dataset, and the y-axis represents the different training strategies. Improvements are highlighted in red, while decreases are shown in blue.}
    \label{fig:fix-ppl}
\end{figure*}

Recent methods \citep{hao2023reasoning, ren2023self} adopt perplexity as a confidence score for LLMs—for example, using the probability of the “A” token to gauge answer confidence in multiple-choice questions. Similarly, we compared this perplexity-based strategy with a fixed 3-shot prompt. As shown in Figure~\ref{fig:fix-ppl}, relying on perplexity markedly degrades model performance on most tasks, which is consistent with prior findings \citep{huang2023large, hong2023closer, he2024advancing}. In other words, perplexity alone is not a sufficiently reliable confidence measure.

\subsection{Performance Across Training–Inference Combinations}
The performance of different training and inference combinations for each dataset is presented in the tables below.

\begin{table*}[!ht]
\caption{Experimental results on the \textbf{Query Intent} dataset (Accuracy \& macro-F1 Score, \%). \colorbox{blue!30}{Blue} highlights the best inference strategy for each training method, while \textbf{bold} denotes the overall best performance across all settings.}
\label{tab:sft_QI}
\centering
\scriptsize
\setlength{\tabcolsep}{0.5pt}
\begin{tabular}{@{}l|ccccc|ccccc|ccccc@{}}
\toprule
\multirow{3}{*}{\textbf{Method}} & \multicolumn{5}{c|}{\textbf{1-shot}} & \multicolumn{5}{c|}{\textbf{3-shot}} & \multicolumn{5}{c}{\textbf{fix\_3\_shot}} \\
\cmidrule(lr){2-6} \cmidrule(lr){7-11} \cmidrule(lr){12-16}
 & fmt-suc & fmt-suc & fmt-suc & overall & overall & fmt-suc & fmt-suc & fmt-suc & overall & overall & fmt-suc & fmt-suc & fmt-suc & overall & overall \\
 & ratio & acc & macro-f1 & acc & macro-f1 & ratio & acc & macro-f1 & acc & macro-f1 & ratio & acc & macro-f1 & acc & macro-f1 \\
\midrule
Zero-shot & 100 & 89.23 & 83.29 & 89.23 & 83.29 & 100 & 91.08 & 84.99 & 91.08 & 84.99 & 100 & 92.09 & 86.14 & 92.09 & 86.14 \\
1-shot & 100 & \cellcolor{blue!30}92.48 & \cellcolor{blue!30}\textbf{87.34} & \cellcolor{blue!30}92.48 & \cellcolor{blue!30}\textbf{87.34} & 100 & 92.47 & 87.20 & 92.47 & 87.20 & 100 & 92.38 & 86.97 & 92.38 & 86.97 \\
3-shot & 100 & 92.29 & 85.98 & 92.29 & 85.98 & 100 & 92.30 & 86.31 & 92.30 & 86.31 & 100 & 92.24 & 85.91 & 92.24 & 85.91 \\
5-shot & 100 & 92.41 & 86.78 & 92.41 & 86.78 & 100 & \cellcolor{blue!30}\textbf{92.52} & 86.95 & \cellcolor{blue!30}\textbf{92.52} & 86.95 & 100 & 92.45 & \cellcolor{blue!30}87.03 & 92.45 & \cellcolor{blue!30}87.03 \\
Definition & 100 & 91.51 & 84.62 & 91.51 & 84.62 & 100 & 91.90 & 85.26 & 91.90 & 85.26 & 100 & 92.19 & 85.62 & 92.19 & 85.62 \\
Numerical & 100 & 0 & 0 & 0 & 0 & 100 & 0 & 0 & 0 & 0 & 100 & 0 & 0 & 0 & 0 \\
Similar-3-shot & 100 & 91.33 & 84.56 & 91.33 & 84.56 & 100 & 91.43 & 85.03 & 91.43 & 85.03 & 100 & 91.47 & 85.19 & 91.47 & 85.19 \\
Fixed-3-shot & 100 & 92.09 & 85.58 & 92.09 & 85.58 & 100 & 92.19 & 85.65 & 92.19 & 85.65 & 100 & \cellcolor{blue!30}92.36 & \cellcolor{blue!30}86.17 & \cellcolor{blue!30}92.36 & \cellcolor{blue!30}86.17 \\
Uncertainty & 100 & 89.82 & 83.71 & 89.82 & 83.71 & 100 & 91.15 & 85.24 & 91.15 & 85.24 & 100 & 92.24 & 86.41 & 92.24 & 86.41 \\
1-shot w/ Def & 100 & \cellcolor{blue!30}92.41 & 78.02 & \cellcolor{blue!30}92.41 & 78.02 & 100 & 92.28 & 77.88 & 92.28 & 77.88 & 100 & 92.29 & 86.35 & 92.29 & 86.35 \\
\midrule
\multirow{3}{*}{\textbf{Method}} & \multicolumn{5}{c|}{\textbf{5\_shot}} & \multicolumn{5}{c|}{\textbf{category\_definition}} & \multicolumn{5}{c}{\textbf{numerical}} \\
\cmidrule(lr){2-6} \cmidrule(lr){7-11} \cmidrule(lr){12-16}
 & fmt-suc & fmt-suc & fmt-suc & overall & overall & fmt-suc & fmt-suc & fmt-suc & overall & overall & fmt-suc & fmt-suc & fmt-suc & overall & overall \\
 & ratio & acc & macro-f1 & acc & macro-f1 & ratio & acc & macro-f1 & acc & macro-f1 & ratio & acc & macro-f1 & acc & macro-f1 \\
\midrule
Zero-shot & 100 & 91.43 & 85.35 & 91.43 & 85.35 & 100 & \cellcolor{blue!30}92.30 & 86.27 & \cellcolor{blue!30}92.30 & 86.27 & 0 & -- & -- & -- & -- \\
1-shot & 100 & 92.43 & 87.12 & 92.43 & 87.12 & 100 & 92.39 & 87.05 & 92.39 & 87.05 & 0 & -- & -- & -- & -- \\
3-shot & 100 & \cellcolor{blue!30}92.44 & \cellcolor{blue!30}86.55 & \cellcolor{blue!30}92.44 & \cellcolor{blue!30}86.55 & 100 & 92.26 & 86.06 & 92.26 & 86.06 & 0 & -- & -- & -- & -- \\
5-shot & 100 & 92.44 & 86.91 & 92.44 & 86.91 & 100 & 92.41 & 86.81 & 92.41 & 86.81 & 0 & -- & -- & -- & -- \\
Definition & 100 & 92.10 & \cellcolor{blue!30}85.91 & 92.10 & \cellcolor{blue!30}85.91 & 100 & \cellcolor{blue!30}92.23 & 85.75 & \cellcolor{blue!30}92.23 & 85.75 & 0 & -- & -- & -- & -- \\
Numerical & 100 & 0 & 0 & 0 & 0 & 100 & 0 & 0 & 0 & 0 & 100 & \cellcolor{blue!30}\textbf{92.52} & \cellcolor{blue!30}86.34 & \cellcolor{blue!30}\textbf{92.52} & \cellcolor{blue!30}86.34 \\
Similar-3-shot & 100 & 91.58 & 85.34 & 91.58 & 85.34 & 100 & 91.60 & 85.36 & 91.60 & 85.36 & 0 & -- & -- & -- & -- \\
Fixed-3-shot & 100 & 92.06 & 85.61 & 92.06 & 85.61 & 100 & 92.14 & 85.58 & 92.14 & 85.58 & 0 & -- & -- & -- & -- \\
Uncertainty & 100 & 91.41 & 85.37 & 91.41 & 85.37 & 100 & 92.29 & 86.03 & 92.29 & 86.03 & 0.11 & 81.82 & 48.72 & 0.09 & 0.05 \\
1-shot w/ Def & 100 & 92.36 & \cellcolor{blue!30}86.57 & 92.36 & \cellcolor{blue!30}86.57 & 100 & 92.26 & 86.37 & 92.26 & 86.37 & 0 & -- & -- & -- & -- \\
\midrule
\multirow{3}{*}{\textbf{Method}} & \multicolumn{5}{c|}{\textbf{similar\_3\_shot}} & \multicolumn{5}{c|}{\textbf{zero\_shot}} & \multicolumn{5}{c}{\textbf{ppl}} \\
\cmidrule(lr){2-6} \cmidrule(lr){7-11} \cmidrule(lr){12-16}
 & fmt-suc & fmt-suc & fmt-suc & overall & overall & fmt-suc & fmt-suc & fmt-suc & overall & overall & fmt-suc & fmt-suc & fmt-suc & overall & overall \\
 & ratio & acc & macro-f1 & acc & macro-f1 & ratio & acc & macro-f1 & acc & macro-f1 & ratio & acc & macro-f1 & acc & macro-f1 \\
\midrule
Zero-shot & 100 & 91.05 & 84.99 & 91.05 & 84.99 & 100 & 92.28 & \cellcolor{blue!30}86.33 & 92.28 & \cellcolor{blue!30}86.33 & 100 & 92.05 & 86.14 & 92.05 & 86.14 \\
1-shot & 100 & 92.41 & 87.20 & 92.41 & 87.20 & 100 & 92.39 & 87.13 & 92.39 & 87.13 & 100 & 92.27 & 86.89 & 92.27 & 86.89 \\
3-shot & 100 & 92.24 & 86.53 & 92.24 & 86.53 & 96.73 & 92.10 & 86.28 & 89.09 & 83.46 & 100 & 92.01 & 86.00 & 92.01 & 86.00 \\
5-shot & 100 & 92.33 & 86.60 & 92.33 & 86.60 & 99.77 & 92.44 & 86.81 & 92.23 & 86.61 & 100 & 92.17 & 86.56 & 92.17 & 86.56 \\
Definition & 100 & 91.40 & 84.87 & 91.40 & 84.87 & 100 & 92.21 & 85.67 & 92.21 & 85.67 & 100 & 92.13 & 85.78 & 92.13 & 85.78 \\
Numerical & 100 & 0 & 0 & 0 & 0 & 100 & 0 & 0 & 0 & 0 & 100 & 56.21 & 26.25 & 56.21 & 26.25 \\
Similar-3-shot & 100 & \cellcolor{blue!30}92.22 & \cellcolor{blue!30}86.40 & \cellcolor{blue!30}92.22 & \cellcolor{blue!30}86.40 & 99.91 & 91.50 & 85.01 & 91.42 & 84.93 & 100 & 91.31 & 84.99 & 91.31 & 84.99 \\
Fixed-3-shot & 100 & 91.62 & 85.63 & 91.62 & 85.63 & 99.87 & 92.09 & 85.51 & 91.97 & 85.40 & 100 & 91.77 & 85.39 & 91.77 & 85.39 \\
Uncertainty & 100 & 91.06 & 85.05 & 91.06 & 85.05 & 100 & \cellcolor{blue!30}92.36 & \cellcolor{blue!30}86.51 & \cellcolor{blue!30}92.36 & \cellcolor{blue!30}86.51 & 100 & 92.27 & 86.10 & 92.27 & 86.10 \\
1-shot w/ Def & 100 & 92.01 & 86.22 & 92.01 & 86.22 & 100 & 92.09 & 67.53 & 92.09 & 67.53 & 100 & 91.92 & 70.37 & 91.92 & 70.37 \\
\bottomrule
\end{tabular}
\end{table*}

\begin{table*}
\caption{Experimental results on the \textbf{Search Correlation} dataset (Accuracy \& macro-F1 Score, \%). \colorbox{blue!30}{Blue} highlights the best inference strategy for each training method, while \textbf{bold} denotes the overall best performance across all settings.}
\label{tab:sft_SC}
\centering
\scriptsize
\setlength{\tabcolsep}{0.5pt}
\begin{tabular}{@{}l|ccccc|ccccc|ccccc@{}}
\toprule
\multirow{3}{*}{\textbf{Method}} & \multicolumn{5}{c|}{\textbf{1-shot}} & \multicolumn{5}{c|}{\textbf{3-shot}} & \multicolumn{5}{c}{\textbf{fix\_3\_shot}} \\
\cmidrule(lr){2-6} \cmidrule(lr){7-11} \cmidrule(lr){12-16}
 & fmt-suc & fmt-suc & fmt-suc & overall & overall & fmt-suc & fmt-suc & fmt-suc & overall & overall & fmt-suc & fmt-suc & fmt-suc & overall & overall \\
 & ratio & acc & macro-f1 & acc & macro-f1 & ratio & acc & macro-f1 & acc & macro-f1 & ratio & acc & macro-f1 & acc & macro-f1 \\
\midrule
Zero-shot & 100 & 63.86 & 56.02 & 63.86 & 56.02 & 100 & 65.30 & 56.16 & 65.30 & 56.16 & 100 & 64.16 & 53.56 & 64.16 & 53.56 \\
1-shot & 100 & \cellcolor{blue!30}67.37 & \cellcolor{blue!30}59.10 & \cellcolor{blue!30}67.37 & \cellcolor{blue!30}59.10 & 100 & 66.43 & 58.19 & 66.43 & 58.19 & 100 & 66.37 & 57.33 & 66.37 & 57.33 \\
3-shot & 100 & 67.17 & 58.45 & 67.17 & 58.45 & 100 & 67.20 & 58.70 & 67.20 & 58.70 & 100 & 67.23 & 58.20 & 67.23 & 58.20 \\
5-shot & 100 & 66.43 & 55.75 & 66.43 & 55.75 & 100 & 66.43 & 56.06 & 66.43 & 56.06 & 100 & 66.27 & 55.51 & 66.27 & 55.51 \\
Definition & 100 & 67.17 & 59.77 & 67.17 & 59.77 & 100 & 66.73 & 59.16 & 66.73 & 59.16 & 100 & 67.10 & 59.15 & 67.10 & 59.15 \\
Numerical & 100 & 0 & 0 & 0 & 0 & 100 & 0 & 0 & 0 & 0 & 100 & 0 & 0 & 0 & 0 \\
Similar-3-shot & 100 & 64.66 & 57.27 & 64.66 & 57.27 & 100 & 64.80 & 57.57 & 64.80 & 57.57 & 100 & 65.27 & 58.01 & 65.27 & 58.01 \\
Fixed-3-shot & 100 & 64.36 & 50.86 & 64.36 & 50.86 & 100 & 64.06 & 51.47 & 64.06 & 51.47 & 100 & 64.26 & 51.02 & 64.26 & 51.02 \\
Uncertainty & 100 & 61.66 & 48.82 & 61.66 & 48.82 & 100 & 61.53 & 47.44 & 61.53 & 47.44 & 100 & 62.06 & 47.17 & 62.06 & 47.17 \\
1-shot w/ Def & 100 & 67.40 & 58.51 & 67.40 & 58.51 & 100 & 66.57 & 57.35 & 66.57 & 57.35 & 100 & 65.70 & 55.47 & 65.70 & 55.47 \\
\midrule
\multirow{3}{*}{\textbf{Method}} & \multicolumn{5}{c|}{\textbf{5\_shot}} & \multicolumn{5}{c|}{\textbf{category\_definition}} & \multicolumn{5}{c}{\textbf{numerical}} \\
\cmidrule(lr){2-6} \cmidrule(lr){7-11} \cmidrule(lr){12-16}
 & fmt-suc & fmt-suc & fmt-suc & overall & overall & fmt-suc & fmt-suc & fmt-suc & overall & overall & fmt-suc & fmt-suc & fmt-suc & overall & overall \\
 & ratio & acc & macro-f1 & acc & macro-f1 & ratio & acc & macro-f1 & acc & macro-f1 & ratio & acc & macro-f1 & acc & macro-f1 \\
\midrule
Zero-shot & 100 & 64.73 & 55.34 & 64.73 & 55.34 & 100 & 66.70 & \cellcolor{blue!30}58.68 & 66.70 & \cellcolor{blue!30}58.68 & 4.14 & 64.52 & 44.25 & 2.67 & 1.83 \\
1-shot & 100 & 66.23 & 47.94 & 66.23 & 47.94 & 100 & 66.57 & 58.19 & 66.57 & 58.19 & 3 & 3.33 & 1.29 & 0.10 & 0.04 \\
3-shot & 100 & \cellcolor{blue!30}67.53 & \cellcolor{blue!30}59.23 & \cellcolor{blue!30}67.53 & \cellcolor{blue!30}59.23 & 100 & 66.87 & 58.52 & 66.87 & 58.52 & 6.14 & 19.57 & 23.63 & 1.20 & 1.45 \\
5-shot & 100 & \cellcolor{blue!30}66.73 & 56.42 & \cellcolor{blue!30}66.73 & 56.42 & 100 & 66.47 & \cellcolor{blue!30}56.54 & 66.47 & \cellcolor{blue!30}56.54 & 49.58 & 56.86 & 45.83 & 28.19 & 22.73 \\
Definition & 100 & 66.17 & 58.35 & 66.17 & 58.35 & 100 & \cellcolor{blue!30}\textbf{68.60} & \cellcolor{blue!30}\textbf{62.25} & \cellcolor{blue!30}\textbf{68.60} & \cellcolor{blue!30}\textbf{62.25} & 0 & -- & -- & -- & -- \\
Numerical & 100 & 0 & 0 & 0 & 0 & 100 & 0 & 0 & 0 & 0 & 100 & \cellcolor{blue!30}64.40 & \cellcolor{blue!30}47.89 & \cellcolor{blue!30}64.40 & \cellcolor{blue!30}47.89 \\
Similar-3-shot & 100 & 65.30 & 58.12 & 65.30 & 58.12 & 100 & 64.80 & 57.40 & 64.80 & 57.40 & 0.13 & 75.00 & 42.86 & 0.10 & 0.06 \\
Fixed-3-shot & 100 & 64.40 & \cellcolor{blue!30}51.80 & 64.40 & \cellcolor{blue!30}51.80 & 100 & 63.23 & 48.88 & 63.23 & 48.88 & 37.64 & 34.22 & 29.29 & 12.88 & 11.02 \\
Uncertainty & 100 & 61.76 & 47.77 & 61.76 & 47.77 & 100 & 62.73 & 48.62 & 62.73 & 48.62 & 0 & -- & -- & -- & -- \\
1-shot w/ Def & 100 & 66.40 & 57.13 & 66.40 & 57.13 & 100 & \cellcolor{blue!30}67.40 & \cellcolor{blue!30}59.07 & \cellcolor{blue!30}67.40 & \cellcolor{blue!30}59.07 & 21.05 & 64.18 & 44.11 & 13.51 & 9.29 \\
\midrule
\multirow{3}{*}{\textbf{Method}} & \multicolumn{5}{c|}{\textbf{similar\_3\_shot}} & \multicolumn{5}{c|}{\textbf{zero\_shot}} & \multicolumn{5}{c}{\textbf{ppl}} \\
\cmidrule(lr){2-6} \cmidrule(lr){7-11} \cmidrule(lr){12-16}
 & fmt-suc & fmt-suc & fmt-suc & overall & overall & fmt-suc & fmt-suc & fmt-suc & overall & overall & fmt-suc & fmt-suc & fmt-suc & overall & overall \\
 & ratio & acc & macro-f1 & acc & macro-f1 & ratio & acc & macro-f1 & acc & macro-f1 & ratio & acc & macro-f1 & acc & macro-f1 \\
\midrule
Zero-shot & 100 & 59.73 & 47.63 & 59.73 & 47.63 & 100 & \cellcolor{blue!30}65.27 & 54.44 & \cellcolor{blue!30}65.27 & 54.44 & 100 & 41.07 & 33.73 & 41.07 & 33.73 \\
1-shot & 100 & 65.23 & 53.24 & 65.23 & 53.24 & 100 & 66.40 & 57.71 & 66.40 & 57.71 & 100 & 31.26 & 23.15 & 31.26 & 23.15 \\
3-shot & 100 & 66.67 & 55.98 & 66.67 & 55.98 & 100 & 67.00 & 58.47 & 67.00 & 58.47 & 100 & 28.43 & 20.29 & 28.43 & 20.29 \\
5-shot & 100 & 65.63 & 52.74 & 65.63 & 52.74 & 100 & 65.80 & 54.10 & 65.80 & 54.10 & 100 & 25.26 & 16.78 & 25.26 & 16.78 \\
Definition & 100 & 65.47 & 54.38 & 65.47 & 54.38 & 100 & 68.20 & 60.99 & 68.20 & 60.99 & 100 & 29.70 & 21.78 & 29.70 & 21.78 \\
Numerical & 100 & 0 & 0 & 0 & 0 & 100 & 0 & 0 & 0 & 0 & 100 & 32.73 & 23.72 & 32.73 & 23.72 \\
Similar-3-shot & 100 & \cellcolor{blue!30}67.63 & \cellcolor{blue!30}60.01 & \cellcolor{blue!30}67.63 & \cellcolor{blue!30}60.01 & 100 & 65.97 & 57.31 & 65.97 & 57.31 & 100 & 25.19 & 17.31 & 25.19 & 17.31 \\
Fixed-3-shot & 100 & \cellcolor{blue!30}64.76 & 50.08 & \cellcolor{blue!30}64.76 & 50.08 & 100 & 63.50 & 48.29 & 63.50 & 48.29 & 100 & 34.70 & 24.71 & 34.70 & 24.71 \\
Uncertainty & 100 & 60.89 & 45.80 & 60.89 & 45.80 & 100 & \cellcolor{blue!30}65.57 & \cellcolor{blue!30}53.18 & \cellcolor{blue!30}65.57 & \cellcolor{blue!30}53.18 & 100 & 38.64 & 28.08 & 38.64 & 28.08 \\
1-shot w/ Def & 100 & 65.53 & 53.36 & 65.53 & 53.36 & 100 & 67.03 & 57.99 & 67.03 & 57.99 & 100 & 29.66 & 21.41 & 29.66 & 21.41 \\
\bottomrule
\end{tabular}
\end{table*}

\begin{table*}
\caption{Experimental results on the \textbf{Query Taxonomy} dataset (Accuracy \& macro-F1 Score, \%). \colorbox{blue!30}{Blue} highlights the best inference strategy for each training method, while \textbf{bold} denotes the overall best performance across all settings.}
\label{tab:sft_QT}
\centering
\scriptsize
\setlength{\tabcolsep}{0.5pt}
\begin{tabular}{@{}l|ccccc|ccccc|ccccc@{}}
\toprule
\multirow{3}{*}{\textbf{Method}} & \multicolumn{5}{c|}{\textbf{1-shot}} & \multicolumn{5}{c|}{\textbf{3-shot}} & \multicolumn{5}{c}{\textbf{fix\_3\_shot}} \\
\cmidrule(lr){2-6} \cmidrule(lr){7-11} \cmidrule(lr){12-16}
 & fmt-suc & fmt-suc & fmt-suc & overall & overall & fmt-suc & fmt-suc & fmt-suc & overall & overall & fmt-suc & fmt-suc & fmt-suc & overall & overall \\
 & ratio & acc & macro-f1 & acc & macro-f1 & ratio & acc & macro-f1 & acc & macro-f1 & ratio & acc & macro-f1 & acc & macro-f1 \\
\midrule
Zero-shot & 100 & 51.53 & 42.6 & 51.53 & 42.6 & 100 & 51.47 & 42.92 & 51.47 & 42.92 & 100 & 51.68 & 42.99 & 51.68 & 42.99 \\
1-shot & 100 & 52.9 & \cellcolor{blue!30}\textbf{44.91} & 52.9 & \cellcolor{blue!30}\textbf{44.91} & 100 & 52.43 & 44.56 & 52.43 & 44.56 & 100 & 52.08 & 44.34 & 52.08 & 44.34 \\
3-shot & 100 & 51.25 & 43.56 & 51.25 & 43.56 & 100 & 52.1 & \cellcolor{blue!30}44.07 & 52.1 & \cellcolor{blue!30}44.07 & 100 & 51.61 & 43.72 & 51.61 & 43.72 \\
5-shot & 100 & 51.31 & 43.44 & 51.31 & 43.44 & 100 & 51.11 & 43.25 & 51.11 & 43.25 & 100 & 50.97 & 43.38 & 50.97 & 43.38 \\
Numerical & 100 & 0 & 0 & 0 & 0 & 100 & 0 & 0 & 0 & 0 & 100 & 0 & 0 & 0 & 0 \\
Similar-3-shot & 100 & 48.56 & 40.75 & 48.56 & 40.75 & 100 & 49.46 & 41.81 & 49.46 & 41.81 & 100 & 48.25 & 41.31 & 48.25 & 41.31 \\
Fixed-3-shot & 100 & 50.74 & 42.49 & 50.74 & 42.49 & 100 & 49.62 & 42.32 & 49.62 & 42.32 & 100 & 50.37 & 42.39 & 50.37 & 42.39 \\
Uncertainty & 100 & 48.13 & 37.42 & 48.13 & 37.42 & 100 & 49.16 & 38.13 & 49.16 & 38.13 & 100 & 48.77 & 37.67 & 48.77 & 37.67 \\
\midrule
\multirow{3}{*}{\textbf{Method}} & \multicolumn{5}{c|}{\textbf{5\_shot}} & \multicolumn{5}{c|}{\textbf{category\_definition}} & \multicolumn{5}{c}{\textbf{numerical}} \\
\cmidrule(lr){2-6} \cmidrule(lr){7-11} \cmidrule(lr){12-16}
 & fmt-suc & fmt-suc & fmt-suc & overall & overall & fmt-suc & fmt-suc & fmt-suc & overall & overall & fmt-suc & fmt-suc & fmt-suc & overall & overall \\
 & ratio & acc & macro-f1 & acc & macro-f1 & ratio & acc & macro-f1 & acc & macro-f1 & ratio & acc & macro-f1 & acc & macro-f1 \\
\midrule
Zero-shot & 100 & 51.3 & 42.77 & 51.3 & 42.77 & -- & -- & -- & -- & -- & 100 & 0 & 0 & 0 & 0 \\
1-shot & 100 & 51.83 & 44.3 & 51.83 & 44.3 & -- & -- & -- & -- & -- & 100 & 0 & 0 & 0 & 0 \\
3-shot & 100 & 51.54 & 43.44 & 51.54 & 43.44 & -- & -- & -- & -- & -- & 56.13 & 0 & 0 & 0 & 0 \\
5-shot & 100 & 51.82 & 43.66 & 51.82 & 43.66 & -- & -- & -- & -- & -- & 93.11 & 0 & 0 & 0 & 0 \\
Numerical & 100 & 0 & 0 & 0 & 0 & -- & -- & -- & -- & -- & 100 & \cellcolor{blue!30}51.26 & \cellcolor{blue!30}42.87 & \cellcolor{blue!30}51.26 & \cellcolor{blue!30}42.87 \\
Similar-3-shot & 100 & 47.99 & 41.23 & 47.99 & 41.23 & -- & -- & -- & -- & -- & 84.55 & 0 & 0 & 0 & 0 \\
Fixed-3-shot & 100 & 50.48 & 42.44 & 50.48 & 42.44 & -- & -- & -- & -- & -- & 55.38 & 0 & 0 & 0 & 0 \\
Uncertainty & 100 & 47.66 & 37.52 & 47.66 & 37.52 & -- & -- & -- & -- & -- & 100 & 0 & 0 & 0 & 0 \\
\midrule
\multirow{3}{*}{\textbf{Method}} & \multicolumn{5}{c|}{\textbf{similar\_3\_shot}} & \multicolumn{5}{c|}{\textbf{zero\_shot}} & \multicolumn{5}{c}{\textbf{ppl}} \\
\cmidrule(lr){2-6} \cmidrule(lr){7-11} \cmidrule(lr){12-16}
 & fmt-suc & fmt-suc & fmt-suc & overall & overall & fmt-suc & fmt-suc & fmt-suc & overall & overall & fmt-suc & fmt-suc & fmt-suc & overall & overall \\
 & ratio & acc & macro-f1 & acc & macro-f1 & ratio & acc & macro-f1 & acc & macro-f1 & ratio & acc & macro-f1 & acc & macro-f1 \\
\midrule
Zero-shot & 100 & \cellcolor{blue!30}53.25 & \cellcolor{blue!30}43.13 & \cellcolor{blue!30}53.25 & \cellcolor{blue!30}43.13 & 100 & 51.43 & 43.1 & 51.43 & 43.1 & -- & -- & -- & -- & -- \\
1-shot & 100 & \cellcolor{blue!30}53.09 & 44.02 & \cellcolor{blue!30}53.09 & 44.02 & 100 & 52.09 & 44.59 & 52.09 & 44.59 & -- & -- & -- & -- & -- \\
3-shot & 100 & \cellcolor{blue!30}53.38 & 43.84 & \cellcolor{blue!30}53.38 & 43.84 & 13.38 & 56.65 & 51.14 & 7.58 & 6.84 & -- & -- & -- & -- & -- \\
5-shot & 100 & \cellcolor{blue!30}53.95 & \cellcolor{blue!30}44.52 & \cellcolor{blue!30}53.95 & \cellcolor{blue!30}44.52 & 52.18 & 58.63 & 51.97 & 30.6 & 27.12 & -- & -- & -- & -- & -- \\
Numerical & 100 & 0 & 0 & 0 & 0 & 100 & 0 & 0 & 0 & 0 & -- & -- & -- & -- & -- \\
Similar-3-shot & 100 & \cellcolor{blue!30}52.99 & \cellcolor{blue!30}44.11 & \cellcolor{blue!30}52.99 & \cellcolor{blue!30}44.11 & 43.95 & 52.54 & 45.77 & 23.09 & 20.11 & -- & -- & -- & -- & -- \\
Fixed-3-shot & 100 & \cellcolor{blue!30}\textbf{54.03} & \cellcolor{blue!30}43.6 & \cellcolor{blue!30}\textbf{54.03} & \cellcolor{blue!30}43.6 & 22.42 & 53.72 & 47.8 & 12.04 & 10.72 & -- & -- & -- & -- & -- \\
Uncertainty & 100 & \cellcolor{blue!30}50.21 & \cellcolor{blue!30}38.35 & \cellcolor{blue!30}50.21 & \cellcolor{blue!30}38.35 & 100 & 48.74 & 38.1 & 48.74 & 38.1 & -- & -- & -- & -- & -- \\
\bottomrule
\end{tabular}
\end{table*}

\begin{table*}
\caption{Experimental results on the \textbf{EC} dataset (Accuracy \& macro-F1 Score, \%). \colorbox{blue!30}{Blue} highlights the best inference strategy for each training method, while \textbf{bold} denotes the overall best performance across all settings.}
\label{tab:sft_EC}
\centering
\scriptsize
\setlength{\tabcolsep}{0.5pt}
\begin{tabular}{@{}l|ccccc|ccccc|ccccc@{}}
\toprule
\multirow{3}{*}{\textbf{Method}} & \multicolumn{5}{c|}{\textbf{1-shot}} & \multicolumn{5}{c|}{\textbf{3-shot}} & \multicolumn{5}{c}{\textbf{fix\_3\_shot}} \\
\cmidrule(lr){2-6} \cmidrule(lr){7-11} \cmidrule(lr){12-16}
 & fmt-suc & fmt-suc & fmt-suc & overall & overall & fmt-suc & fmt-suc & fmt-suc & overall & overall & fmt-suc & fmt-suc & fmt-suc & overall & overall \\
 & ratio & acc & macro-f1 & acc & macro-f1 & ratio & acc & macro-f1 & acc & macro-f1 & ratio & acc & macro-f1 & acc & macro-f1 \\
\midrule
Zero-shot & 100 & 83.30 & 78.97 & 83.30 & 78.97 & 100 & 92.60 & 88.73 & 92.60 & 88.73 & 100 & 91.95 & 88.53 & 91.95 & 88.53 \\
1-shot & 100 & 92.95 & 88.97 & 92.95 & 88.97 & 100 & 92.80 & 88.76 & 92.80 & 88.76 & 100 & 92.75 & 88.62 & 92.75 & 88.62 \\
3-shot & 100 & 93.40 & 89.10 & 93.40 & 89.10 & 100 & 93.25 & 88.71 & 93.25 & 88.71 & 100 & 93.20 & 88.46 & 93.20 & 88.46 \\
5-shot & 100 & 93.90 & 89.05 & 93.90 & 89.05 & 100 & 93.75 & 88.99 & 93.75 & 88.99 & 100 & 93.65 & 88.94 & 93.65 & 88.94 \\
Definition & 100 & 78.85 & 73.68 & 78.85 & 73.68 & 100 & 92.35 & 87.19 & 92.35 & 87.19 & 100 & 92.00 & 87.04 & 92.00 & 87.04 \\
Numerical & 100 & 38.80 & 1.67 & 38.80 & 1.67 & 100 & 56.55 & 2.58 & 56.55 & 2.58 & 100 & 43.05 & 1.71 & 43.05 & 1.71 \\
Similar-3-shot & 100 & 93.20 & 88.43 & 93.20 & 88.43 & 100 & 93.25 & 88.55 & 93.25 & 88.55 & 100 & 93.60 & 89.08 & 93.60 & 89.08 \\
Fixed-3-shot & 100 & 93.60 & 89.56 & 93.60 & 89.56 & 100 & 93.45 & 89.04 & 93.45 & 89.04 & 100 & \cellcolor{blue!30}93.80 & \cellcolor{blue!30}89.92 & \cellcolor{blue!30}93.80 & \cellcolor{blue!30}89.92 \\
Uncertainty & 100 & 77.25 & 73.31 & 77.25 & 73.31 & 100 & 91.45 & 88.06 & 91.45 & 88.06 & 100 & 92.50 & 89.07 & 92.50 & 89.07 \\
1-shot w/ Def & 100 & 93.60 & 89.43 & 93.60 & 89.43 & 100 & 93.80 & 89.70 & 93.80 & 89.70 & 100 & 93.35 & 89.65 & 93.35 & 89.65 \\
\midrule
\multirow{3}{*}{\textbf{Method}} & \multicolumn{5}{c|}{\textbf{5\_shot}} & \multicolumn{5}{c|}{\textbf{category\_definition}} & \multicolumn{5}{c}{\textbf{numerical}} \\
\cmidrule(lr){2-6} \cmidrule(lr){7-11} \cmidrule(lr){12-16}
 & fmt-suc & fmt-suc & fmt-suc & overall & overall & fmt-suc & fmt-suc & fmt-suc & overall & overall & fmt-suc & fmt-suc & fmt-suc & overall & overall \\
 & ratio & acc & macro-f1 & acc & macro-f1 & ratio & acc & macro-f1 & acc & macro-f1 & ratio & acc & macro-f1 & acc & macro-f1 \\
\midrule
Zero-shot & 100 & 93.50 & 89.92 & 93.50 & 89.92 & 100 & \cellcolor{blue!30}93.75 & \cellcolor{blue!30}\textbf{90.31} & \cellcolor{blue!30}93.75 & \cellcolor{blue!30}\textbf{90.31} & 0 & -- & -- & -- & -- \\
1-shot & 100 & 92.85 & 88.62 & 92.85 & 88.62 & 63.60 & 94.50 & 90.98 & 60.10 & 57.86 & 0 & -- & -- & -- & -- \\
3-shot & 100 & 93.20 & 88.63 & 93.20 & 88.63 & 19.55 & 94.88 & 95.94 & 18.55 & 18.76 & 0 & -- & -- & -- & -- \\
5-shot & 100 & \cellcolor{blue!30}\textbf{94.15} & \cellcolor{blue!30}89.83 & \cellcolor{blue!30}\textbf{94.15} & \cellcolor{blue!30}89.83 & 99.85 & 93.84 & 89.48 & 93.70 & 89.34 & 0 & -- & -- & -- & -- \\
Definition & 100 & 92.85 & 88.05 & 92.85 & 88.05 & 100 & 93.15 & 88.04 & 93.15 & 88.04 & 0 & -- & -- & -- & -- \\
Numerical & 100 & 63.50 & 3.12 & 63.50 & 3.12 & 100 & 62.45 & 5.70 & 62.45 & 5.70 & 100 & \cellcolor{blue!30}93.65 & \cellcolor{blue!30}89.93 & \cellcolor{blue!30}93.65 & \cellcolor{blue!30}89.93 \\
Similar-3-shot & 100 & 93.55 & 88.71 & 93.55 & 88.71 & 3.05 & 78.69 & 89.79 & 2.40 & 2.74 & 6.25 & 87.20 & 73.43 & 5.45 & 4.59 \\
Fixed-3-shot & 100 & 93.35 & 89.06 & 93.35 & 89.06 & 85.25 & 93.61 & 89.11 & 79.80 & 75.97 & 0 & -- & -- & -- & -- \\
Uncertainty & 100 & 92.75 & 89.27 & 92.75 & 89.27 & 87.30 & 92.50 & 89.50 & 80.75 & 78.14 & 0 & -- & -- & -- & -- \\
1-shot w/ Def & 100 & 93.35 & 88.88 & 93.35 & 88.88 & 100 & \cellcolor{blue!30}93.80 & \cellcolor{blue!30}89.95 & \cellcolor{blue!30}93.80 & \cellcolor{blue!30}89.95 & 0 & -- & -- & -- & -- \\
\midrule
\multirow{3}{*}{\textbf{Method}} & \multicolumn{5}{c|}{\textbf{similar\_3\_shot}} & \multicolumn{5}{c|}{\textbf{zero\_shot}} & \multicolumn{5}{c}{\textbf{ppl}} \\
\cmidrule(lr){2-6} \cmidrule(lr){7-11} \cmidrule(lr){12-16}
 & fmt-suc & fmt-suc & fmt-suc & overall & overall & fmt-suc & fmt-suc & fmt-suc & overall & overall & fmt-suc & fmt-suc & fmt-suc & overall & overall \\
 & ratio & acc & macro-f1 & acc & macro-f1 & ratio & acc & macro-f1 & acc & macro-f1 & ratio & acc & macro-f1 & acc & macro-f1 \\
\midrule
Zero-shot & 100 & 92.35 & 88.73 & 92.35 & 88.73 & 100 & 93.70 & 90.17 & 93.70 & 90.17 & 100 & 93.65 & 89.96 & 93.65 & 89.96 \\
1-shot & 100 & 93.00 & 88.93 & 93.00 & 88.93 & 100 & \cellcolor{blue!30}93.15 & \cellcolor{blue!30}89.45 & \cellcolor{blue!30}93.15 & \cellcolor{blue!30}89.45 & 100 & 91.05 & 87.30 & 91.05 & 87.30 \\
3-shot & 100 & 93.05 & 88.27 & 93.05 & 88.27 & 99.95 & \cellcolor{blue!30}93.50 & 89.17 & \cellcolor{blue!30}93.45 & 89.13 & 100 & 93.20 & \cellcolor{blue!30}89.39 & 93.20 & \cellcolor{blue!30}89.39 \\
5-shot & 100 & 93.60 & 89.17 & 93.60 & 89.17 & 100 & 93.70 & 89.06 & 93.70 & 89.06 & 100 & 93.55 & 89.19 & 93.55 & 89.19 \\
Definition & 100 & 92.10 & 87.12 & 92.10 & 87.12 & 100 & \cellcolor{blue!30}93.30 & \cellcolor{blue!30}88.31 & \cellcolor{blue!30}93.30 & \cellcolor{blue!30}88.31 & 100 & 92.90 & 88.30 & 92.90 & 88.30 \\
Numerical & 100 & 63.20 & 2.65 & 63.20 & 2.65 & 100 & 55.35 & 2.88 & 55.35 & 2.88 & 100 & 92.45 & 88.74 & 92.45 & 88.74 \\
Similar-3-shot & 100 & \cellcolor{blue!30}93.90 & \cellcolor{blue!30}89.44 & \cellcolor{blue!30}93.90 & \cellcolor{blue!30}89.44 & 39.65 & 94.70 & 92.81 & 37.55 & 36.80 & 100 & 91.95 & 87.81 & 91.95 & 87.81 \\
Fixed-3-shot & 100 & 93.25 & 89.33 & 93.25 & 89.33 & 100 & 93.20 & 89.14 & 93.20 & 89.14 & 100 & 92.90 & 88.92 & 92.90 & 88.92 \\
Uncertainty & 100 & 91.90 & 88.37 & 91.90 & 88.37 & 100 & \cellcolor{blue!30}93.55 & \cellcolor{blue!30}90.01 & \cellcolor{blue!30}93.55 & \cellcolor{blue!30}90.01 & 100 & 93.05 & 89.49 & 93.05 & 89.49 \\
1-shot w/ Def & 100 & 93.00 & 88.09 & 93.00 & 88.09 & 100 & 93.80 & 89.81 & 93.80 & 89.81 & 100 & 92.50 & 88.93 & 92.50 & 88.93 \\
\bottomrule
\end{tabular}
\end{table*}

\begin{table*}
\caption{Experimental results on the \textbf{IFLYTEK} dataset (Accuracy \& macro-F1 Score, \%). \colorbox{blue!30}{Blue} highlights the best inference strategy for each training method, while \textbf{bold} denotes the overall best performance across all settings.}
\label{tab:sft_IT}
\centering
\scriptsize
\setlength{\tabcolsep}{0.5pt}
\begin{tabular}{@{}l|ccccc|ccccc|ccccc@{}}
\toprule
\multirow{3}{*}{\textbf{Method}} & \multicolumn{5}{c|}{\textbf{1-shot}} & \multicolumn{5}{c|}{\textbf{3-shot}} & \multicolumn{5}{c}{\textbf{fix\_3\_shot}} \\
\cmidrule(lr){2-6} \cmidrule(lr){7-11} \cmidrule(lr){12-16}
 & fmt-suc & fmt-suc & fmt-suc & overall & overall & fmt-suc & fmt-suc & fmt-suc & overall & overall & fmt-suc & fmt-suc & fmt-suc & overall & overall \\
 & ratio & acc & macro-f1 & acc & macro-f1 & ratio & acc & macro-f1 & acc & macro-f1 & ratio & acc & macro-f1 & acc & macro-f1 \\
\midrule
Zero-shot & 100 & 60.95 & 44.95 & 60.95 & 44.95 & 100 & 60.95 & 45.33 & 60.95 & 45.33 & 100 & 61.06 & 45.32 & 61.06 & 45.32 \\
1-shot & 100 & 63.22 & 46.75 & 63.22 & 46.75 & 100 & \cellcolor{blue!30}63.33 & \cellcolor{blue!30}47.15 & \cellcolor{blue!30}63.33 & \cellcolor{blue!30}47.15 & 100 & 63.22 & 47.15 & 63.22 & 47.15 \\
3-shot & 100 & \cellcolor{blue!30}62.91 & \cellcolor{blue!30}\textbf{48.51} & \cellcolor{blue!30}62.91 & \cellcolor{blue!30}\textbf{48.51} & 100 & 62.33 & 47.33 & 62.33 & 47.33 & 100 & 62.91 & 47.22 & 62.91 & 47.22 \\
5-shot & 100 & 61.91 & 44.60 & 61.91 & 44.60 & 100 & 61.75 & 44.05 & 61.75 & 44.05 & 100 & 61.75 & 43.26 & 61.75 & 43.26 \\
Definition & 100 & 63.41 & 44.86 & 63.41 & 44.86 & 100 & \cellcolor{blue!30}63.64 & 44.41 & \cellcolor{blue!30}63.64 & 44.41 & 100 & 63.26 & 44.49 & 63.26 & 44.49 \\
Numerical & 100 & 53.67 & 22.11 & 53.67 & 22.11 & 100 & 54.83 & 24.27 & 54.83 & 24.27 & 100 & 54.41 & 22.55 & 54.41 & 22.55 \\
Similar-3-shot & 100 & 62.10 & 46.43 & 62.10 & 46.43 & 100 & 62.68 & 47.19 & 62.68 & 47.19 & 100 & \cellcolor{blue!30}62.83 & 47.69 & \cellcolor{blue!30}62.83 & 47.69 \\
Fixed-3-shot & 100 & 63.49 & 45.88 & 63.49 & 45.88 & 100 & 63.37 & 45.22 & 63.37 & 45.22 & 100 & \cellcolor{blue!30}63.52 & 45.26 & \cellcolor{blue!30}63.52 & 45.26 \\
Uncertainty & 100 & 63.33 & 46.91 & 63.33 & 46.91 & 100 & 63.26 & 46.73 & 63.26 & 46.73 & 100 & 63.06 & 46.45 & 63.06 & 46.45 \\
1-shot w/ Def & 100 & 63.06 & 46.47 & 63.06 & 46.47 & 100 & 63.06 & 46.62 & 63.06 & 46.62 & 100 & \cellcolor{blue!30}63.37 & \cellcolor{blue!30}47.47 & \cellcolor{blue!30}63.37 & \cellcolor{blue!30}47.47 \\
\midrule
\multirow{3}{*}{\textbf{Method}} & \multicolumn{5}{c|}{\textbf{5\_shot}} & \multicolumn{5}{c|}{\textbf{category\_definition}} & \multicolumn{5}{c}{\textbf{numerical}} \\
\cmidrule(lr){2-6} \cmidrule(lr){7-11} \cmidrule(lr){12-16}
 & fmt-suc & fmt-suc & fmt-suc & overall & overall & fmt-suc & fmt-suc & fmt-suc & overall & overall & fmt-suc & fmt-suc & fmt-suc & overall & overall \\
 & ratio & acc & macro-f1 & acc & macro-f1 & ratio & acc & macro-f1 & acc & macro-f1 & ratio & acc & macro-f1 & acc & macro-f1 \\
\midrule
Zero-shot & 100 & 61.02 & 45.27 & 61.02 & 45.27 & 100 & 61.56 & \cellcolor{blue!30}46.93 & 61.56 & \cellcolor{blue!30}46.93 & 0 & -- & -- & -- & -- \\
1-shot & 100 & 63.18 & 46.45 & 63.18 & 46.45 & 99.65 & 63.17 & 47.91 & 62.95 & 47.75 & 0.15 & 75.00 & 42.86 & 0.12 & 0.07 \\
3-shot & 100 & 62.41 & 47.34 & 62.41 & 47.34 & 98.46 & 63.03 & 48.74 & 62.06 & 47.99 & 2.12 & 41.82 & 15.85 & 0.88 & 0.34 \\
5-shot & 100 & 61.99 & 44.01 & 61.99 & 44.01 & 100 & \cellcolor{blue!30}62.52 & \cellcolor{blue!30}44.93 & \cellcolor{blue!30}62.52 & \cellcolor{blue!30}44.93 & 0.15 & 50.00 & 22.22 & 0.08 & 0.03 \\
Definition & 100 & 63.45 & 44.35 & 63.45 & 44.35 & 100 & 62.83 & 44.90 & 62.83 & 44.90 & 0 & -- & -- & -- & -- \\
Numerical & 100 & 53.79 & 22.34 & 53.79 & 22.34 & 100 & 50.29 & 27.23 & 50.29 & 27.23 & 100 & \cellcolor{blue!30}62.29 & \cellcolor{blue!30}46.29 & \cellcolor{blue!30}62.29 & \cellcolor{blue!30}46.29 \\
Similar-3-shot & 100 & 62.64 & 47.29 & 62.64 & 47.29 & 100 & 62.37 & \cellcolor{blue!30}47.72 & 62.37 & \cellcolor{blue!30}47.72 & 0.27 & 42.86 & 12.00 & 0.12 & 0.03 \\
Fixed-3-shot & 100 & 63.29 & 45.08 & 63.29 & 45.08 & 100 & 63.10 & \cellcolor{blue!30}46.43 & 63.10 & \cellcolor{blue!30}46.43 & 0.08 & 0.00 & 0.00 & 0.00 & 0.00 \\
Uncertainty & 100 & 63.45 & 46.42 & 63.45 & 46.42 & 100 & \cellcolor{blue!30}\textbf{63.76} & \cellcolor{blue!30}47.85 & \cellcolor{blue!30}\textbf{63.76} & \cellcolor{blue!30}47.85 & 0 & -- & -- & -- & -- \\
1-shot w/ Def & 100 & 63.26 & 46.46 & 63.26 & 46.46 & 100 & 62.52 & 46.36 & 62.52 & 46.36 & 0 & -- & -- & -- & -- \\
\midrule
\multirow{3}{*}{\textbf{Method}} & \multicolumn{5}{c|}{\textbf{similar\_3\_shot}} & \multicolumn{5}{c|}{\textbf{zero\_shot}} & \multicolumn{5}{c}{\textbf{ppl}} \\
\cmidrule(lr){2-6} \cmidrule(lr){7-11} \cmidrule(lr){12-16}
 & fmt-suc & fmt-suc & fmt-suc & overall & overall & fmt-suc & fmt-suc & fmt-suc & overall & overall & fmt-suc & fmt-suc & fmt-suc & overall & overall \\
 & ratio & acc & macro-f1 & acc & macro-f1 & ratio & acc & macro-f1 & acc & macro-f1 & ratio & acc & macro-f1 & acc & macro-f1 \\
\midrule
Zero-shot & 100 & \cellcolor{blue!30}62.83 & 46.69 & \cellcolor{blue!30}62.83 & 46.69 & 100 & 61.18 & 45.42 & 61.18 & 45.42 & 100 & 38.21 & 31.67 & 38.21 & 31.67 \\
1-shot & 100 & 62.06 & 45.15 & 62.06 & 45.15 & 99.35 & 63.44 & 46.39 & 63.02 & 46.09 & 100 & 39.28 & 30.76 & 39.28 & 30.76 \\
3-shot & 100 & 62.29 & 46.16 & 62.29 & 46.16 & 97.81 & 62.90 & 47.41 & 61.52 & 46.37 & 100 & 42.17 & 33.51 & 42.17 & 33.51 \\
5-shot & 100 & 61.83 & 43.97 & 61.83 & 43.97 & 100 & 61.87 & 43.96 & 61.87 & 43.96 & 100 & 41.98 & 31.24 & 41.98 & 31.24 \\
Definition & 100 & 62.60 & 43.88 & 62.60 & 43.88 & 100 & 63.45 & \cellcolor{blue!30}45.41 & 63.45 & \cellcolor{blue!30}45.41 & 100 & 28.43 & 23.69 & 28.43 & 23.69 \\
Numerical & 100 & 58.64 & 29.57 & 58.64 & 29.57 & 100 & 50.98 & 20.52 & 50.98 & 20.52 & 100 & 46.44 & 38.71 & 46.44 & 38.71 \\
Similar-3-shot & 100 & 62.45 & 46.37 & 62.45 & 46.37 & 100 & 62.49 & 47.45 & 62.49 & 47.45 & 100 & 38.51 & 29.83 & 38.51 & 29.83 \\
Fixed-3-shot & 100 & 63.14 & 44.78 & 63.14 & 44.78 & 100 & 62.72 & 43.99 & 62.72 & 43.99 & 100 & 40.40 & 30.23 & 40.40 & 30.23 \\
Uncertainty & 100 & 63.45 & 47.15 & 63.45 & 47.15 & 100 & 63.64 & 46.61 & 63.64 & 46.61 & 100 & 46.25 & 35.77 & 46.25 & 35.77 \\
1-shot w/ Def & 100 & 62.75 & 46.72 & 62.75 & 46.72 & 100 & 63.18 & 46.73 & 63.18 & 46.73 & 100 & 19.51 & 15.56 & 19.51 & 15.56 \\
\bottomrule
\end{tabular}
\end{table*}

\begin{table*}
\caption{Experimental results on the \textbf{TNEWS} dataset (Accuracy \& macro-F1 Score, \%). \colorbox{blue!30}{Blue} highlights the best inference strategy for each training method, while \textbf{bold} denotes the overall best performance across all settings.}
\label{tab:sft_TN}
\centering
\scriptsize
\setlength{\tabcolsep}{0.5pt}
\begin{tabular}{@{}l|ccccc|ccccc|ccccc@{}}
\toprule
\multirow{3}{*}{\textbf{Method}} & \multicolumn{5}{c|}{\textbf{1-shot}} & \multicolumn{5}{c|}{\textbf{3-shot}} & \multicolumn{5}{c}{\textbf{fix\_3\_shot}} \\
\cmidrule(lr){2-6} \cmidrule(lr){7-11} \cmidrule(lr){12-16}
 & fmt-suc & fmt-suc & fmt-suc & overall & overall & fmt-suc & fmt-suc & fmt-suc & overall & overall & fmt-suc & fmt-suc & fmt-suc & overall & overall \\
 & ratio & acc & macro-f1 & acc & macro-f1 & ratio & acc & macro-f1 & acc & macro-f1 & ratio & acc & macro-f1 & acc & macro-f1 \\
\midrule
Zero-shot & 100 & 61.31 & 58.96 & 61.31 & 58.96 & 100 & 61.31 & 58.70 & 61.31 & 58.70 & 100 & 61.33 & 58.75 & 61.33 & 58.75 \\
1-shot & 100 & 61.23 & 59.98 & 61.23 & 59.98 & 100 & 61.45 & 60.08 & 61.45 & 60.08 & 100 & 61.41 & 60.15 & 61.41 & 60.15 \\
3-shot & 100 & 61.07 & 59.56 & 61.07 & 59.56 & 100 & 61.21 & 60.09 & 61.21 & 60.09 & 100 & 61.30 & 60.25 & 61.30 & 60.25 \\
5-shot & 100 & 61.73 & 57.66 & 61.73 & 57.66 & 100 & 61.48 & 57.40 & 61.48 & 57.40 & 100 & 61.48 & 57.72 & 61.48 & 57.72 \\
Definition & 100 & 60.48 & 59.00 & 60.48 & 59.00 & 100 & 60.84 & 59.56 & 60.84 & 59.56 & 100 & 60.56 & 59.27 & 60.56 & 59.27 \\
Numerical & 100 & 54.12 & 2.65 & 54.12 & 2.65 & 100 & 55.07 & 3.89 & 55.07 & 3.89 & 100 & 55.86 & 5.05 & 55.86 & 5.05 \\
Similar-3-shot & 100 & 60.60 & 59.02 & 60.60 & 59.02 & 100 & 60.59 & 59.07 & 60.59 & 59.07 & 100 & 60.47 & 59.20 & 60.47 & 59.20 \\
Fixed-3-shot & 100 & 61.11 & 57.92 & 61.11 & 57.92 & 100 & 61.07 & 58.45 & 61.07 & 58.45 & 100 & 60.91 & 58.66 & 60.91 & 58.66 \\
Uncertainty & 100 & 60.54 & 56.71 & 60.54 & 56.71 & 100 & 60.77 & 56.86 & 60.77 & 56.86 & 100 & 60.84 & 56.97 & 60.84 & 56.97 \\
1-shot w/ Def & 100 & 61.09 & 59.92 & 61.09 & 59.92 & 100 & 61.32 & 60.09 & 61.32 & 60.09 & 100 & 61.31 & 60.02 & 61.31 & 60.02 \\
\midrule
\multirow{3}{*}{\textbf{Method}} & \multicolumn{5}{c|}{\textbf{5\_shot}} & \multicolumn{5}{c|}{\textbf{category\_definition}} & \multicolumn{5}{c}{\textbf{numerical}} \\
\cmidrule(lr){2-6} \cmidrule(lr){7-11} \cmidrule(lr){12-16}
 & fmt-suc & fmt-suc & fmt-suc & overall & overall & fmt-suc & fmt-suc & fmt-suc & overall & overall & fmt-suc & fmt-suc & fmt-suc & overall & overall \\
 & ratio & acc & macro-f1 & acc & macro-f1 & ratio & acc & macro-f1 & acc & macro-f1 & ratio & acc & macro-f1 & acc & macro-f1 \\
\midrule
Zero-shot & 100 & 61.52 & 59.12 & 61.52 & 59.12 & 100 & 61.52 & 58.98 & 61.52 & 58.98 & 0 & -- & -- & -- & -- \\
1-shot & 100 & 61.35 & 60.08 & 61.35 & 60.08 & 100 & 61.29 & 59.92 & 61.29 & 59.92 & 0 & -- & -- & -- & -- \\
3-shot & 100 & 61.39 & 60.22 & 61.39 & 60.22 & 100 & 61.16 & 59.98 & 61.16 & 59.98 & 0 & -- & -- & -- & -- \\
5-shot & 100 & 61.82 & 58.02 & 61.82 & 58.02 & 100 & 61.76 & 57.61 & 61.76 & 57.61 & 0 & -- & -- & -- & -- \\
Definition & 100 & 60.89 & \cellcolor{blue!30}59.73 & 60.89 & \cellcolor{blue!30}59.73 & 100 & 60.63 & 59.38 & 60.63 & 59.38 & 0 & -- & -- & -- & -- \\
Numerical & 100 & 55.10 & 4.64 & 55.10 & 4.64 & 100 & 31.56 & 0.43 & 31.56 & 0.43 & 100 & \cellcolor{blue!30}61.24 & \cellcolor{blue!30}57.09 & \cellcolor{blue!30}61.24 & \cellcolor{blue!30}57.09 \\
Similar-3-shot & 100 & 60.59 & 59.17 & 60.59 & 59.17 & 100 & 60.67 & 58.89 & 60.67 & 58.89 & 0 & -- & -- & -- & -- \\
Fixed-3-shot & 100 & 61.01 & 58.40 & 61.01 & 58.40 & 100 & 60.80 & 57.19 & 60.80 & 57.19 & 0 & -- & -- & -- & -- \\
Uncertainty & 100 & 60.80 & 56.96 & 60.80 & 56.96 & 100 & 60.73 & 56.85 & 60.73 & 56.85 & 0 & -- & -- & -- & -- \\
1-shot w/ Def & 100 & 61.35 & 60.19 & 61.35 & 60.19 & 100 & 61.22 & 59.85 & 61.22 & 59.85 & 0 & -- & -- & -- & -- \\
\midrule
\multirow{3}{*}{\textbf{Method}} & \multicolumn{5}{c|}{\textbf{similar\_3\_shot}} & \multicolumn{5}{c|}{\textbf{zero\_shot}} & \multicolumn{5}{c}{\textbf{ppl}} \\
\cmidrule(lr){2-6} \cmidrule(lr){7-11} \cmidrule(lr){12-16}
 & fmt-suc & fmt-suc & fmt-suc & overall & overall & fmt-suc & fmt-suc & fmt-suc & overall & overall & fmt-suc & fmt-suc & fmt-suc & overall & overall \\
 & ratio & acc & macro-f1 & acc & macro-f1 & ratio & acc & macro-f1 & acc & macro-f1 & ratio & acc & macro-f1 & acc & macro-f1 \\
\midrule
Zero-shot & 100 & \cellcolor{blue!30}\textbf{63.30} & \cellcolor{blue!30}\textbf{61.31} & \cellcolor{blue!30}\textbf{63.30} & \cellcolor{blue!30}\textbf{61.31} & 100 & 60.83 & 59.39 & 60.83 & 59.39 & 100 & 45.68 & 35.99 & 45.68 & 35.99 \\
1-shot & 100 & \cellcolor{blue!30}61.98 & \cellcolor{blue!30}60.49 & \cellcolor{blue!30}61.98 & \cellcolor{blue!30}60.49 & 100 & 61.10 & 59.75 & 61.10 & 59.75 & 100 & 49.46 & 41.59 & 49.46 & 41.59 \\
3-shot & 100 & \cellcolor{blue!30}62.06 & \cellcolor{blue!30}60.53 & \cellcolor{blue!30}62.06 & \cellcolor{blue!30}60.53 & 100 & 61.15 & 60.07 & 61.15 & 60.07 & 100 & 43.84 & 36.87 & 43.84 & 36.87 \\
5-shot & 100 & \cellcolor{blue!30}62.54 & \cellcolor{blue!30}58.41 & \cellcolor{blue!30}62.54 & \cellcolor{blue!30}58.41 & 100 & 61.94 & 57.85 & 61.94 & 57.85 & 100 & 47.88 & 39.06 & 47.88 & 39.06 \\
Definition & 100 & \cellcolor{blue!30}61.37 & 59.26 & \cellcolor{blue!30}61.37 & 59.26 & 100 & 60.63 & 59.66 & 60.63 & 59.66 & 100 & 46.21 & 38.56 & 46.21 & 38.56 \\
Numerical & 100 & 59.64 & 12.99 & 59.64 & 12.99 & 100 & 39.16 & 0.54 & 39.16 & 0.54 & 100 & 53.68 & 50.42 & 53.68 & 50.42 \\
Similar-3-shot & 100 & \cellcolor{blue!30}\textbf{63.30}& \cellcolor{blue!30}\textbf{61.31} & \cellcolor{blue!30}\textbf{63.30} & \cellcolor{blue!30}\textbf{61.31} & 100 & 60.83 & 59.39 & 60.83 & 59.39 & 100 & 45.68 & 35.99 & 45.68 & 35.99 \\
Fixed-3-shot & 100 & \cellcolor{blue!30}62.25 & \cellcolor{blue!30}58.94 & \cellcolor{blue!30}62.25 & \cellcolor{blue!30}58.94 & 100 & 60.96 & 58.37 & 60.96 & 58.37 & 100 & 49.46 & 40.44 & 49.46 & 40.44 \\
Uncertainty & 100 & \cellcolor{blue!30}61.97 & \cellcolor{blue!30}58.15 & \cellcolor{blue!30}61.97 & \cellcolor{blue!30}58.15 & 100 & 61.12 & 57.32 & 61.12 & 57.32 & 100 & 49.35 & 41.01 & 49.35 & 41.01 \\
1-shot w/ Def & 100 & \cellcolor{blue!30}62.20 & \cellcolor{blue!30}60.65 & \cellcolor{blue!30}62.20 & \cellcolor{blue!30}60.65 & 100 & 61.30 & 59.90 & 61.30 & 59.90 & 100 & 49.89 & 41.75 & 49.89 & 41.75 \\
\bottomrule
\end{tabular}
\end{table*}

\clearpage
\section{Packing Results}


A common optimization technique in the pre-training stage of LLMs is \textbf{packing}, where multiple training samples are concatenated into a single sequence to improve computational efficiency. When applied to SFT for classification tasks, packing introduces two effects: (i) it increases the effective batch size and context length, and (ii) it allows samples within a packed sequence to attend to preceding samples—referred to as contaminated attention. We hypothesize that this second effect may mimic the behavior of ICL training.

To test this hypothesis, we conducted experiments on seven datasets under three conditions: (i) no packing, (ii) packing, and (iii) packing with attention mask to prevent cross-sample contamination. 


\begin{table*}[!ht]
\caption{Performance of packing strategies across all datasets. Accuracy and macro-F1 (\%) are reported for 1-shot, 3-shot, 5-shot, and zero-shot settings. Cells shaded in green denote cases where standard packing outperforms neat packing, whereas yellow shading indicates the opposite. \textbf{Bold} numbers mark the best results in each column.}

\label{tab:pack_combined}
\centering
\scriptsize
\setlength{\tabcolsep}{4pt}
\label{tab:packing}
\renewcommand{\arraystretch}{0.85}
\begin{tabular}{@{}l|c|cc|cc|cc|cc@{}}
\toprule
\multirow{2}{*}{\textbf{Dataset}} & \multirow{2}{*}{\textbf{Method}} & \multicolumn{2}{c|}{\textbf{1-shot}} & \multicolumn{2}{c|}{\textbf{3-shot}} & \multicolumn{2}{c|}{\textbf{5-shot}} & \multicolumn{2}{c}{\textbf{zero-shot}} \\
\cmidrule(lr){3-4} \cmidrule(lr){5-6} \cmidrule(lr){7-8} \cmidrule(lr){9-10}
& & Acc & Macro-F1 & Acc & Macro-F1 & Acc & Macro-F1 & Acc & Macro-F1 \\
\midrule
\multirow{3}{*}{QI} 
& No Packing & 89.23 & 83.29 & 91.08 & 84.99 & 91.43 & 85.35 & \textbf{92.28} & \textbf{86.33} \\
& Packing & 90.93 & \cellcolor{green!30}83.69 & \cellcolor{green!30}91.30 & \cellcolor{green!30}84.65 & \cellcolor{green!30}91.29 & \cellcolor{green!30}84.85 & 91.20 & 84.28 \\
& Neat Packing & \cellcolor{yellow!30}91.00 & 79.38 & 91.29 & 80.01 & 91.15 & 83.99 & \cellcolor{yellow!30}91.58 & \cellcolor{yellow!30}84.94 \\
\midrule
\multirow{3}{*}{SC} 
& No Packing & 63.86 & 56.02 & 65.30 & 56.16 & 64.73 & 55.34 & \textbf{67.43} & \textbf{58.64} \\
& Packing & \cellcolor{green!30}64.90 & \cellcolor{green!30}51.97 & \cellcolor{green!30}64.76 & \cellcolor{green!30}51.37 & \cellcolor{green!30}65.10 & \cellcolor{green!30}51.96 & 65.07 & 51.47 \\
& Neat Packing & 61.93 & 51.20 & 62.16 & 49.36 & 61.63 & 48.12 & \cellcolor{yellow!30}65.27 & \cellcolor{yellow!30}54.44 \\
\midrule
\multirow{3}{*}{QT} 
& No Packing & 51.53 & 42.60 & 51.47 & 42.92 & 51.30 & 42.77 & 51.43 & 43.10 \\
& Packing & 46.99 & 41.50 & 47.94 & 42.21 & 49.38 & \cellcolor{green!30}\textbf{43.38} & 50.10 & 41.79 \\
& Neat Packing & \cellcolor{yellow!30}51.19 & \cellcolor{yellow!30}42.24 & \cellcolor{yellow!30}\textbf{51.82} & \cellcolor{yellow!30}42.71 & \cellcolor{yellow!30}50.24 & 42.49 & \cellcolor{yellow!30}51.32 & \cellcolor{yellow!30}42.76 \\
\midrule
\multirow{3}{*}{EC} 
& No Packing & 83.30 & 78.97 & 92.60 & 88.73 & 93.50 & 89.92 & \textbf{93.70} & \textbf{90.17} \\
& Packing & \cellcolor{green!30}91.55 & \cellcolor{green!30}87.94 & \cellcolor{green!30}91.20 & \cellcolor{green!30}86.80 & \cellcolor{green!30}91.60 & \cellcolor{green!30}87.03 & 91.95 & 87.05 \\
& Neat Packing & 76.75 & 70.92 & 89.55 & 84.73 & 90.85 & 85.91 & \cellcolor{yellow!30}92.85 & \cellcolor{yellow!30}88.70 \\
\midrule
\multirow{3}{*}{EIC} 
& No Packing & 83.43 & 82.40 & 83.39 & 82.67 & \textbf{84.04} & \textbf{82.93} & 82.74 & 81.73 \\
& Packing & 81.66 & 79.20 & 81.96 & 79.62 & 82.74 & 81.05 & 81.62 & 74.95 \\
& Neat Packing & \cellcolor{yellow!30}82.61 & \cellcolor{yellow!30}80.80 & \cellcolor{yellow!30}83.52 & \cellcolor{yellow!30}82.66 & \cellcolor{yellow!30}83.22 & \cellcolor{yellow!30}82.09 & \cellcolor{yellow!30}83.39 & \cellcolor{yellow!30}82.77 \\
\midrule
\multirow{3}{*}{IT} 
& No Packing & 61.31 & 44.95 & 60.95 & 45.33 & 61.02 & 45.27 & 61.18 & 45.42 \\
& Packing & 63.41 & 47.14 & \cellcolor{green!30}63.83 & \cellcolor{green!30}48.15 & \cellcolor{green!30}\textbf{64.06} & \cellcolor{green!30}47.56 & 63.95 & 48.41 \\
& Neat Packing & \cellcolor{yellow!30}63.49 & \cellcolor{yellow!30}47.50 & 62.75 & 46.91 & 63.18 & 47.24 & \cellcolor{yellow!30}63.99 & \cellcolor{yellow!30}\textbf{48.99} \\
\midrule
\multirow{3}{*}{TN} 
& No Packing & 61.31 & 58.96 & 61.31 & 58.70 & 61.52 & 59.12 & \textbf{61.71} & \textbf{59.51} \\
& Packing & 60.74 & 56.62 & \cellcolor{green!30}60.70 & 56.46 & 60.58 & 56.36 & 60.62 & 56.66 \\
& Neat Packing & \cellcolor{yellow!30}60.86 & \cellcolor{yellow!30}58.33 & 60.65 & \cellcolor{yellow!30}57.41 & \cellcolor{yellow!30}60.84 & \cellcolor{yellow!30}57.42 & \cellcolor{yellow!30}61.40 & \cellcolor{yellow!30}58.58 \\
\bottomrule
\end{tabular}
\end{table*}

From the results in Table \ref{tab:packing}, on QI, SC, EC, IT datasets, we observe that \textbf{neat packing yields higher zero-shot accuracy, while standard packing achieves better few-shot performance}—providing empirical support for our hypothesis that cross-sample attention mimics in-context learning.
Moreover, the no-packing setting yields the best performance on five of the seven datasets, specifically the QI, SC, EC, EIC, and TN.

\end{document}